\documentclass[]{l4dc2022}

\usepackage{times}
\usepackage{hyperref}
\usepackage{url}
\usepackage{multirow}
\usepackage{graphicx}
\usepackage{mathtools}
\usepackage[utf8]{inputenc}

\usepackage[toc,page,header]{appendix}
\usepackage{minitoc}

\usepackage{amsmath,amsfonts,bm}

\def\eqref#1{equation~\ref{#1}}
\def\1{\bm{1}}

\DeclareMathAlphabet{\mathsfit}{\encodingdefault}{\sfdefault}{m}{sl}
\SetMathAlphabet{\mathsfit}{bold}{\encodingdefault}{\sfdefault}{bx}{n}

\def\gA{{\mathcal{A}}}

\def\gG{{\mathcal{G}}}

\def\gM{{\mathcal{M}}}

\def\gS{{\mathcal{S}}}

\def\gX{{\mathcal{X}}}

\renewcommand \thepart{}
\renewcommand \partname{}

\title{Can Foundation Models Perform Zero-Shot \\ Task Specification For Robot Manipulation?}

\author{%
  \Name{Yuchen Cui} \Email{yuchencui@utexas.edu} \\
  \Name{Scott Niekum} \Email{sniekum@cs.utexas.edu}\\
  \addr The University of Texas at Austin
  \AND
 \Name{Abhinav Gupta} \Email{gabhinav@fb.com }\\
 \Name{Vikash Kumar} \Email{vikashplus@fb.com}\\
 \Name{Aravind Rajeswaran} \Email{aravraj@fb.com }\\
 \addr Facebook AI Research 
}

\newcommand{\printfnsymbol}[1]{%
  \textsuperscript{\@fnsymbol{#1}}%
}

\begin{document}

\maketitle

\begin{abstract}%
Task specification is at the core of programming autonomous robots. A low-effort modality for task specification is critical for engagement of non-expert end-users and ultimate adoption of personalized robot agents.
A widely studied approach to task specification is through goals, using either  compact state vectors or goal images from the same robot scene. The former is hard to interpret for non-experts and necessitates detailed state estimation and scene understanding. The latter requires the generation of desired goal image, which often requires a human to complete the task, defeating the purpose of having autonomous robots.
In this work, we explore alternate and more general forms of goal specification that are expected to be easier for humans to specify and use such as images obtained from the internet, hand sketches that provide a visual description of the desired task, or simple language descriptions.
As a preliminary step towards this, we investigate the capabilities of large scale pre-trained models (foundation models) for zero-shot goal specification, and find 
promising results in a collection of simulated robot manipulation tasks and real-world datasets. 
Project webpage: \url{https://sites.google.com/view/zestproject}

\end{abstract}


\doparttoc
\faketableofcontents

\section{Introduction}
\label{sec:intro}

% talk about how this relates to common sense model
Robots are gradually entering our homes to help automate aspects of our everyday life. End users of such technologies are likely to have personalized requirements and preferences. A low-effort and intuitive modality of communication is needed to allow non-expert users to program robots and customize them to user needs. This requires robots to be equipped with a form of ``common sense'' that is grounded in human-centric experiences and understanding of the world. Recent advances in computer vision~\citep{Krizhevsky2012ImageNetCW} and natural language processing~(NLP)~\citep{Devlin2019BERT, Radford2019GPT} have enabled machine learning models to make sense of images and text using large-scale internet datasets, that comprise in part, data generated by humans in human-centric environments. Of particular interest are ``foundation models''~\citep{bommasani2021opportunities} -- deep neural network models trained on massive internet datasets -- that have powered impressive advances in downstream vision and NLP tasks. In this backdrop, our work explores if such foundation models can also enable task specification for robotics and advance embodied intelligence. 

\begin{figure}[t]
    \centering
    \includegraphics[width=\textwidth]{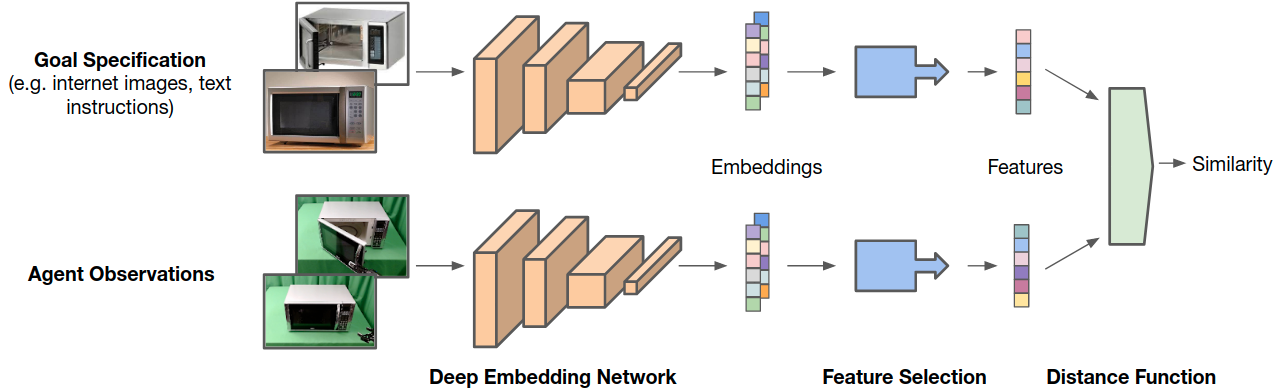}
    \caption{Overview of our proposed \textbf{Ze}ro \textbf{S}hot \textbf{T}ask-specification (ZeST) framework. The observation and goal specifications are first embedded (e.g. using a foundation model). We compute features of these embedding, and subsequently measure the similarity of these features. A high similarity between agent observation and goal implies the specification has been achieved.}
    \label{fig:teaser}
\end{figure}

% para on task/goal specification
Traditional approaches to program robots, both at training and deployment times, involve the use of reward functions, task demonstrations, or goal specification. Among these, goal specification is particularly appealing from an end user viewpoint since goals are easier to provide compared to full demonstrations and also require minimal expertise compared to scripting detailed reward functions that are prone to exploitation~\citep{amodei2016concrete}. Existing goal-conditioned approaches use either compact state space goal vectors or RGB images from the same robot scene. The former is often hard to interpret and requires detailed state estimation, which is difficult outside lab settings. The latter forces a human expert to perform the task to generate the necessary goal image, thereby defeating the purpose of an autonomous robot.
We study two alternate forms of goal conditioning to overcome the aforementioned difficulties. 

The first approach involves goals or instructions in natural language, such as \textit{``open the microwave''}. To be successful, the robot must understand grounding and semantics -- it must be capable of understanding what a microwave looks like and also be capable of differentiating between open and closed microwaves. The second approach involves using goal images from a different scene, such as images from the internet. For example, to instruct a robot to open a microwave, we can provide the robot an image of an open microwave obtained from the internet. To be successful, the robot must have domain adaptation capabilities and form correspondences between the state of the microwave in the goal image and current scene. 
The goal of our work is to study if foundation models are useful for such types of task specification, particularly in the zero-shot regime where they have been very successful in traditional vision and NLP tasks.

\paragraph{Our Contributions:} In this work, (1) We introduce a \textit{framework} for studying foundation models for zero-shot task specification (ZeST). See Figure~\ref{fig:teaser} for an overview.
(2) We evaluate the effectiveness of ZeST for enabling zero-shot \textit{policy execution} through a set of goal selection tasks; 
and (3)~We evaluate ZeST for \textit{policy learning} in offline reinforcement learning.
We find that ZeST is quite effective in zero-shot goal-selection and provides a 14-fold increase in performance over a random guessing baseline. In offline RL, we find that using ZeST scores as a proxy for the reward function enables the learning of policies that perform better than a behavior cloning baseline.

\pagebreak
\section{Background and Related Work}
\label{sec:background}

We consider environments that take the form of a \textbf{High-Dimensional Markov Decision Process (MDPs)}, described by the tuple  $\gM = \langle \gS, \gX, \gA, P, R, d_0, \gamma \rangle$. Informally, this setting has been widely studied and was characterized more formally by \citet{Du2019ProvablyER} as ``Block MDPs''. Here, $\gS$ denotes a compact state space that is not directly observable by the agent, $\gX$ denotes the high-dimensional observation space that contains sufficient information to uniquely recover the underlying state, and $\gA$ is the action space. 
For example, the observation space can be (multi-view) camera images while the state space can be object poses. 
The underlying transition dynamics and reward function are described in the state space as $P(s'|s,a)$ and $R(s)$. Additionally, $d_0$ is the starting state distribution and $\gamma \in [0, 1)$ is the discount factor. Since the observation can be mapped uniquely to the state, an equivalent MDP can be constructed in the observation space as well.
A trajectory $\tau = \{(X_0,a_0),(X_1,a_1),...,(X_H,a_H)\}$ is a sequence of observation-action pairs of length $H$. 
% The return of a trajectory is $\sum_{t=0}^H[\gamma^t R(s_t)]$. 
A policy, $\pi(a_t|X_t)$ maps from observations to a probability distribution over actions. 
The objective is to learn a policy that maximizes the long term reward, i.e. $\max_{\pi} \mathbb{E} \left[ \sum_{t=0}^\infty \gamma^t R(s_t) \right]$ where the expectation is under $s_0 \sim d_0$, $a_t \sim \pi(\cdot|X_t)$ and $s_{t+1} \sim P(\cdot|s_t, a_t)$.

\paragraph{Goal-Conditioned Policies}
In the standard RL formulation, an agent learns a policy for a single task (reward function). Goal-conditioning allows the agent to perform multiple tasks by conditioning on different goals from a goal space $G \in \gG$. The policy $\pi(a_t|X_t,G)$ and reward function $R(s_t, G)$ can be made a function of the goal, thereby enabling multi-task learning. We can interpret goal-conditioning as constructing a broader MDP with an augmented state and observation spaces $\gS \cup \gG$ and $\gX \cup \gG$. 
The idea of goal-conditioning has been extensively studied. Prior work has considered goals either in the compact state space~\citep{kaelbling1993learning, andrychowicz2017hindsight,fu2018variational,gupta2019relay}, or in high-dimensional image space, but from the same scene as the agent~\citep{nair2018visual, nair2019hierarchical, singh2019end,johns2021coarse}. Compact state vectors are hard to interpret, depend on the scene, and require detailed state estimation. Image based goals are human-interpretable, but requires another agent (typically a human) to perform the task first to generate the image for conditioning. In this work, we propose the use of off-domain images and/or language instructions to reduce the task specification burden on users. Our method can also be adapted to work with existing goal-conditioned learning algorithms. 

\paragraph{Foundation Models and Applications}
A major advancement in modern deep learning is the emergence of representation and transfer learning. Models trained on generic datasets from the internet are capable of learning representations that transfer successfully to a plethora of downstream applications. This observation has been widespread since the seminal works of AlexNet~\citep{Krizhevsky2012ImageNetCW} and ResNet~\citep{he2016deep}.
Recent advances in self-supervised learning have led to further advancements in visual~\citep{he2020momentum}, language~\citep{Devlin2019BERT, Radford2019GPT}, and multi-modal~\citep{radford2021learning} representation models that have widely impacted downstream applications in vision~\citep{tian2020rethinking, conde2021clip, patashnik2021styleclip, frans2021clipdraw} 
and NLP~\citep{lin2021pretrained, bugliarello2021multimodal,izacard2020leveraging}, and for this reason have been touted as ``foundation models''~\citep{bommasani2021opportunities}. However, the use of foundation models for control and embodied intelligence is relatively new and under-explored. CLIPort~\citep{shridhar2021cliport} combines CLIP embedding with transporter network to learn language-conditioned robot manipulation policies from demonstrations. The concurrent works of \citet{khandelwal2021simple} and \citet{Parisi2022PVR}, study the use CLIP and other self-supervised representation networks as a perception module for control tasks and observe they outperform traditional ImageNet-pretrained backbones.

\section{ZeST: Using Foundation Models for Task Specification}
\label{sec:method}

\begin{figure}[b!]
    \centering
    \includegraphics[width=0.95\textwidth]{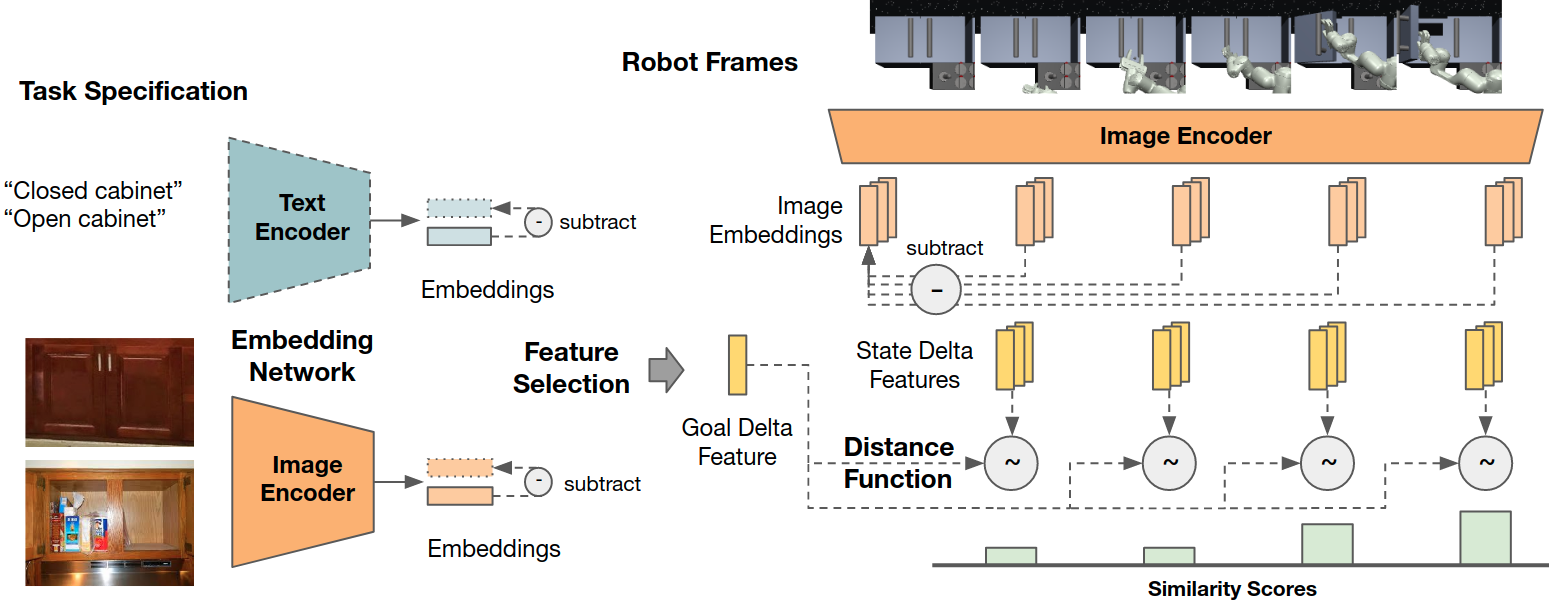}
    \caption{An instantiation of the ZeST framework with delta features. We observe that similarity of the observation (top row robot frames) with the task specification (open cabinet) increases as the robot executes a successful trajectory.}
    \label{fig:zest_instantiation}
\end{figure}

As outlined in Section~\ref{sec:background}, current approaches to goal-conditioning are limited due to requirement of state-vector based goals or goal images from the same scene as agent. 
In this work, we introduce ZeST, a framework for studying more generic and intuitive forms of goal-specification like off-domain images (e.g. those found on the internet) and language instructions. To succeed with this type of goal specification, the agent must capture ``common sense'' that is grounded in human-centric experiences and understanding of the world. Motivated by the advancement in foundation models~\citep{bommasani2021opportunities} that were trained using massive amounts of data from the internet, which contain in part data generated by humans in human-centric environments, we specifically study if foundation models can enable these generic forms of goal-conditioning.
In this work, we also focus on \textbf{zero-shot} goal-specification, where the agent must be capable of performing tasks when simply presented with a user-specified (off-domain) goal. We do not assume any human annotations or supervised pairings between prior experiences of the agent and goal specifications, which is representative of real deployment scenarios. Moreover, foundation models have demonstrated impressive zero-shot learning results~\citep{radford2021learning, Brown2020GPT3}.

At its core, the ZeST framework provides a way to measure the ``similarity'' between the image observation of the agent and a user-specified goal, with high similarity implying the goal specification is satisfied. The most similar image in the replay buffer of the agent can be used for goal-conditioning. The similarity can also be used as a proxy for reward signal in case of policy learning. While the approach of measuring similarity between agent's observation and goal has been explored in prior work, for instance by using a learned classifier~\citep{Pinto2016SupersizingSL, fu2018variational, Eysenbach2021CLearningLT}, they have all required goal images from the same scene as the agent. The overall ZeST framework is described in Figure~\ref{fig:teaser}, and consists of three main components: (1)~embedding network; (2)~feature selector; (3)~feature distance function. Each of the three modules can be instantiated according to properties and constraints of the downstream task.

For the \textbf{embedding network}, we use pre-trained foundation models for embedding both the goal specification and agent observation. While any foundation model can be used, in our experimental evaluation, we focus on ImageNet-supervised ResNet50~\citep{he2016deep}, ImageNet-trained MoCo~\citep{he2020momentum}, and CLIP~\citep{radford2021learning}. Since CLIP is a multi-modal embedding network, it also provides a mechanism for specifying tasks through language descriptions.
Details of the three candidate embedding networks can be found in the Appendix.
The \textbf{feature selection} module further featurizes the embedding. 
In this work, we consider two simple featurizations: 1) \underline{raw features} that use the raw embeddings themselves, and 2) \underline{delta features} that use the difference in embeddings between the desired and initial configurations, and analogously for the agent, the difference in embeddings between current observation and initial observations.
The raw features are useful to directly compare the specified goal with the current observation. With delta features, the user could specify both the goal state and an initial/current state in order to indicate a desired change of an object's state. For example, a user would provide the images of both a closed microwave and open microwave, which serves to better specify the task by describing the high-level action of opening the microwave. For the \textbf{similarity metric}, we explore the use of cosine similarity and L2 distance, both of which have been used in prior work involving representation learning. In summary, the similarity between a pair of observations $(X_t, X_0)$ and goal-specification $(G_0, G_f)$ is written as:
\begin{align}
    \bm{\phi}^{\textrm{raw}} \left( X_t, G_f \right) & := \bm{\alpha} \left( \psi(X_t), \psi(G_f) \right), \\
    \bm{\phi}^{\textrm{delta}} \left( (X_t, X_0), (G_f, G_0) \right) & := \bm{\alpha} \left(  \left( \psi(X_t) - \psi(X_0) \right), \left( \psi(G_f) - \psi(G_0) \right) \right)
\end{align}
where $\psi(\cdot)$ is the embedding function (foundation model) and $\bm{\alpha}$ is the similarity metric (cosine similarity, L2 distance etc.).
Figure~\ref{fig:zest_instantiation} illustrates an instantiation of ZeST with delta features.

We emphasize here that ZeST provides an abstraction to study generic forms of goal conditioning, and the use of various foundation models for the same. While we explore a number of design choices as outlined above and in the Appendix, there are many additional possibilities that all conform to our broad framework. An exhaustive study of all such combinations is beyond the scope of any single work, and we hope that our work inspires further work in the study of goal-conditioning and foundation models for embodied intelligence.

\section{Experimental Tasks and Evaluation Metrics}

In this section, we outline three different tasks that span both policy deployment and policy learning. For each task, we also outline different variations and evaluation metrics.

\subsection{Goal Selection Task}
Given a dataset of experience $D$, off-domain goal-specification $(G_0, G_f)$, and initial/context observation $X_0$, the task for the agent is to select an observation $X_f \in D$ that satisfies the goal-specification in context of $X_0$.
The dataset of experience, $D= \{\tau_1, \tau_2, \ldots \tau_N \}$, in general consists of observation-action trajectories. We remain agnostic to the source of dataset, which could have been obtained through teleoperation or autonomous execution various policies. The task for the agent is to find the image in the dataset that has maximum similarity to the goal-specification in the context of initial image $X_0$. Let $\mu$ denote the number of observations in the dataset that satisfy the goal specification. We evaluate the performance according to two metrics:

\begin{enumerate}
    \item {\bf Top-N Success Rate:} We rank each observation in the dataset according to the similarity measure, and compute the fraction of top $N\leq \mu$ ranked samples that satisfies the goal specification. This metric allows us to study if a high-similarity can be used to retrieve observations close to the goal, and ultimately enable downstream goal-conditioned policy execution.
    
    \item {\bf Dataset Total Variation:} While top-N success rate allows us to evaluate model prediction for high-similarity observations, it is agnostic to how the model behaves for low-similarity observations (i.e. non-goal states). Towards obtaining a broad evaluation metric for similarity scores across the entire dataset, we consider the dataset total variation measure. Let $\bm{\phi_i}$ denote the similarity score of observation $X_i$, and let $I(X_i)$ be an indicator function to denote whether $X_i$ satisfies the goal specification. Then, the dataset total variation is given by:
    \begin{equation}
        \label{eq:dataset_total_variation}
        \mathrm{DTV} := \frac{1}{\left\Vert D \right\Vert}\sum_{X_i\in D} \left| \frac{1}{\mu} I(X_i) - \frac{\bm{\phi_i}}{\sum_{j} \bm{\phi_j}} \right|
    \end{equation}
\end{enumerate}

In our experiments, we average the results over multiple choices of $X_0$ and goal-specification $(G_0, G_f)$ to quantify the overall performance of the similarity measure.

\subsection{High-Level Action Selection Task}

The agent is presented with a dataset of experience $D= \{\tau_1, \tau_2, \ldots \tau_N \}$, where each trajectory consists of an observation-action sequence. The agent is also presented with goal specification $(G_0, G_f)$. Let $\Delta g = \psi(G_f)-\psi(G_0)$ be the goal-feature and $\Delta a_i^k = \psi(I^k_i) - \psi(I^k_0)$ be the high-level action feature for timestep $i$ in trajectory $k$ ($I^k_0$ is the initial image in trajectory $k$).
The task for the agent is to find a pair of images in the dataset that has maximum similarity to the high-level action (i.e. transformation) in $\Delta g$. The action selection task is useful when a user need to specify an action but do not have access to images of the particular object of interest. For example, they could use images of ``opening a cabinet door" to specify ``opening microwave", since both share the same high-level action semantics of ``opening''. For this task, we identify semantically equivalent actions among trajectories and evaluate model performance with Top-N success rate.

% \subsection{Reinforcement Learning} 
% The task specified by $G_0$ and $G_f$, a robot manipulate an object in its own environment (specifically the Franka kitchen domain) that matches the object in $G_0$ and $G_f$. The similarity score output by our proposed method is directly used as the reward for each state/observation. A policy gradient algorithm~\citep{rajeswaran17towards, rajeswaran18learning} is employed for learning from the predicted reward signal. The performance of the learned agents is then evaluated with (hand designed) ground truth return.

\subsection{Offline Reinforcement Learning} 
The task is specified by $G_0$ and $G_f$, and a robot must manipulate an object in its own environment that matches the object in $G_0$ and $G_f$. A set of sub-optimal trajectories are provided for learning consistent with the offline RL setting. \textit{Behavioral cloning} directly learns a supervised model that maps state to action, and thus simply mimics the behavior in the dataset without any improvement. We propose to evaluate ZeST by using an offline RL algorithm to learn the policy using the ZeST similarity scores as a proxy for the reward function. If we observe an improvement over the behavior cloning baseline, then we may conclude that ZeST scores function as a reasonable reward proxy for policy learning.

\section{Experimental Results and Discussion}
\label{sec:experiments}

To test the effectiveness of ZeST, we evaluate performance of various instantiations of ZeST for enabling policy execution and learning. In our experiments, we consider the following design choices for components outlined in Section~\ref{sec:method}.
\begin{enumerate}
    \itemsep0em
    \item {\bf Embedding Foundation Models:} ImageNet-supervised ResNet50~\citep{he2016deep}, ImageNet-trained MoCo~\citep{he2020momentum}, and CLIP~\citep{radford2021learning}. We consider these models primarily due to wide adoption in computer vision, with CLIP being a natural choice for encoding both visual and language goals due to its multi-modal training.
    \item {\bf Features:} Raw embedding and delta feature variants as specified in Section~\ref{sec:method}.
    \item {\bf Feature distance:} Cosine similarity and L2 distance.
\end{enumerate}

\subsection{How does ZeST perform in the Goal Selection Task? How do various design choices impact performance?}

\begin{figure}[t]
    \centering
    \includegraphics[width=0.45\linewidth]{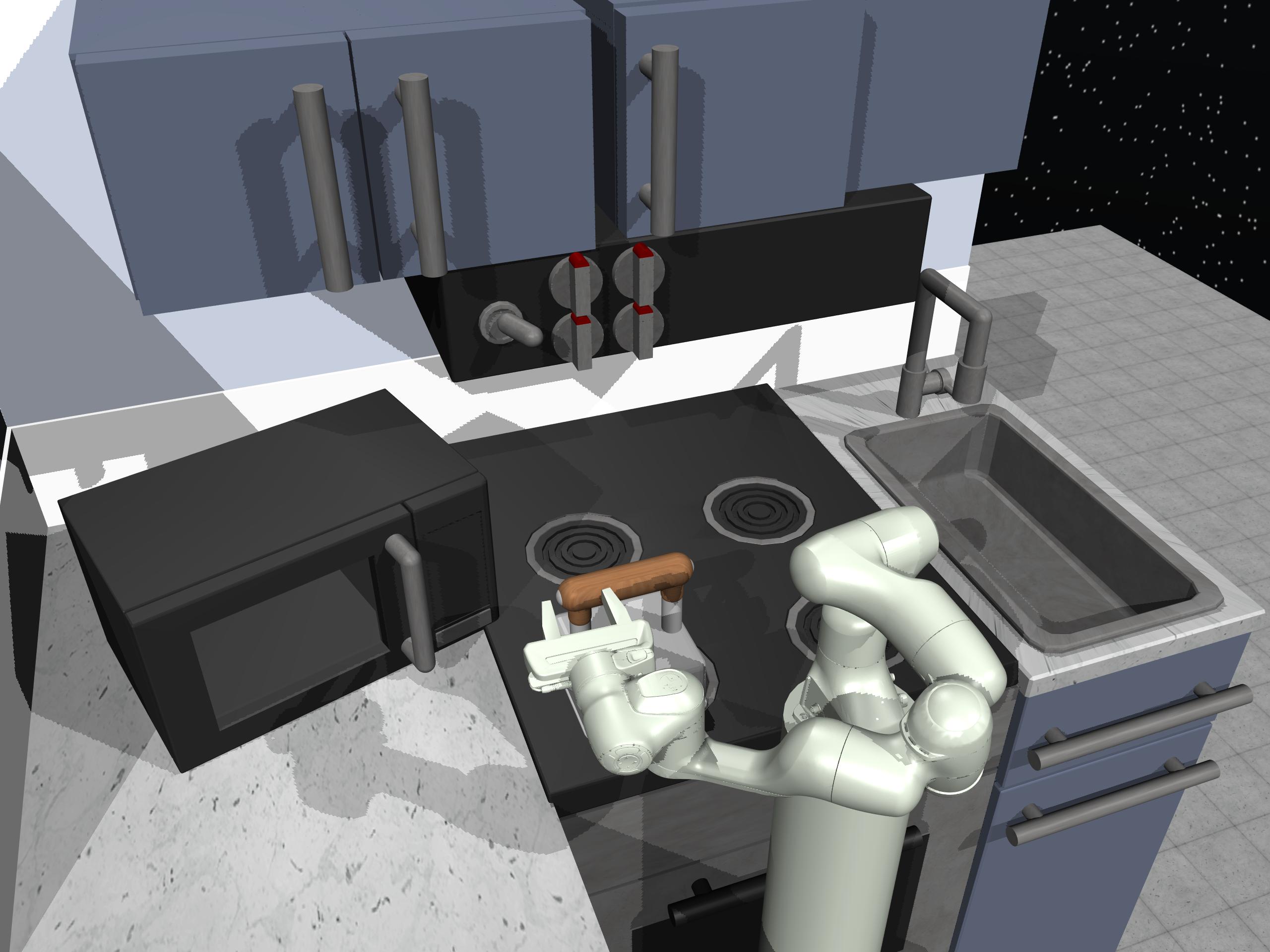}
    \caption{\small Visualization of the Franka Kitchen domain from \citet{gupta2019relay}.}
    \label{views}
\end{figure}
\begin{figure}[b!]
    \centering
    \includegraphics[width=0.95\linewidth]{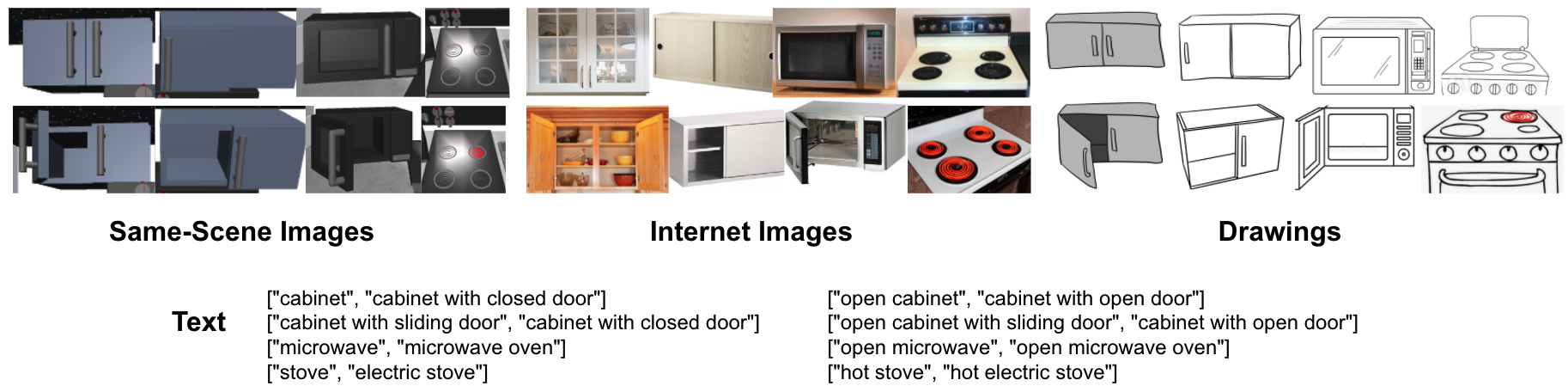}
    \vspace{-0.3cm}
    \caption{\small Samples of different task specification modalities in the Visual Task Dataset~\citep{cui2021dataset}.}
    \label{fig:datasets}
\end{figure}

To evaluate the performance of ZeST, we first consider an experience dataset in the Franka Kitchen domain \citep{gupta2019relay}. The dataset consists of five different manipulation tasks: turning the top/bottom burner knob on, opening the microwave, opening the hinge door, and opening the sliding door. The experience dataset is generated by rolling out randomized expert policies that were trained using a policy gradient algorithm~\citep{rajeswaran17towards, rajeswaran18learning}. Uniform noise of varying levels were added to the actions to create a diverse dataset that contains both successful and failed trajectories.
We evaluate the performance of ZeST using different forms of goal-specification: (1) same-scene images -- i.e. goal images from the Franka Kitchen scene itself; (2) off-domain images from the internet; (3) hand-sketches and drawings; (4) and instructions in natural language. The internet image and drawings dataset are part of the Visual Task Dataset~\citep{cui2021dataset}, and Figure~\ref{fig:datasets} provides a few representative examples from this dataset.

\begin{figure}[t!]
    \centering
    \includegraphics[width=0.85\linewidth]{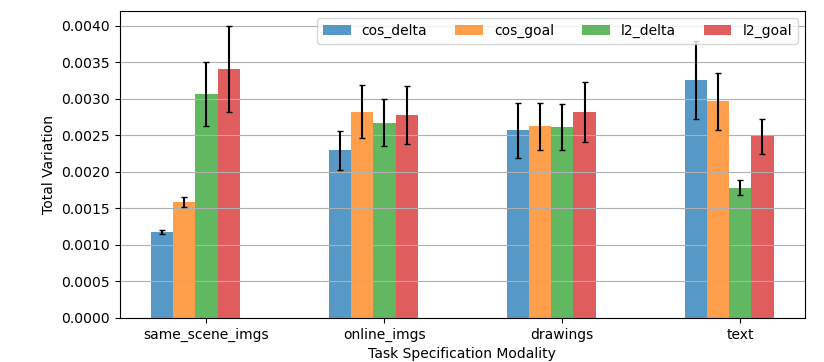}
    \vspace{-0.3cm}
    \caption{Avg.~total variation (with std.~error) across tasks in Franka kitchen domain under different task specification modality for candidate similarity computation methods across embedding models. Cosine similarity of delta features outperforms other methods.}
    \label{fig:total_variance}
    \vspace{-0.4cm}
\end{figure}

We first evaluate the choice of features and feature distance function. To do so, we compute the dataset total variation (Eq.~\ref{eq:dataset_total_variation}), averaged across different embedding models. 
The results are summarized in Figure~\ref{fig:total_variance}. We find that when goals are specified through images, the use of delta features and cosine similarity leads to the best results. In case of language goals, we find that using delta features and L2 distance leads to the lowest total variation. 
Based on these results, for the remainder of the paper, we primarily focus on the best performing design choices.

We consider two different variations for evaluation: (1) Narrow dataset (ND), in which the experience dataset contains trajectories related to only a single manipulation task. For example, the narrow dataset would only contain (suboptimal) trajectories where the Franka robot attempts to manipulate the microwave. (2) Diverse dataset (DD), which is the entire experience dataset comprising of a diverse set of trajectories related to all the five manipulation tasks. The ND scenario is simpler and serves to evaluate if ZeST can perform goal-selection in the context of a single manipulation task. This can also be useful for policy learning similar to prior works that utilize a goal classifier as a reward function~\citep{fu2018variational, Eysenbach2021CLearningLT}. The DD scenario is harder and is also representative of goal-selection for policy execution in open-ended environments. 

\begin{figure}
    \centering
    \hspace{-0.2cm} \includegraphics[width=0.65\linewidth]{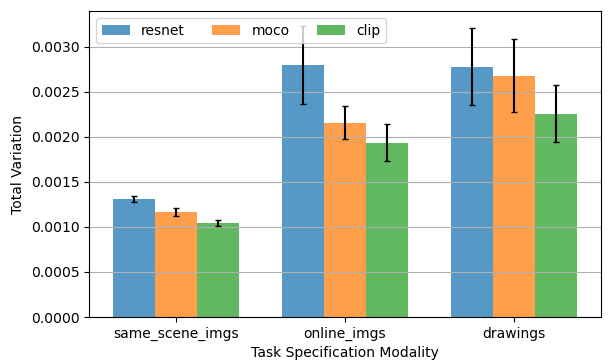}
    \vspace{-0.35cm}
    \caption{Avg.~total variation for three embedding models in Franka Kitchen domain with \textbf{Cos+Delta}.}
    \label{fig:embedding_model_dtv}
\end{figure}

\begin{figure}
    \centering
    \includegraphics[width=0.5\textwidth]{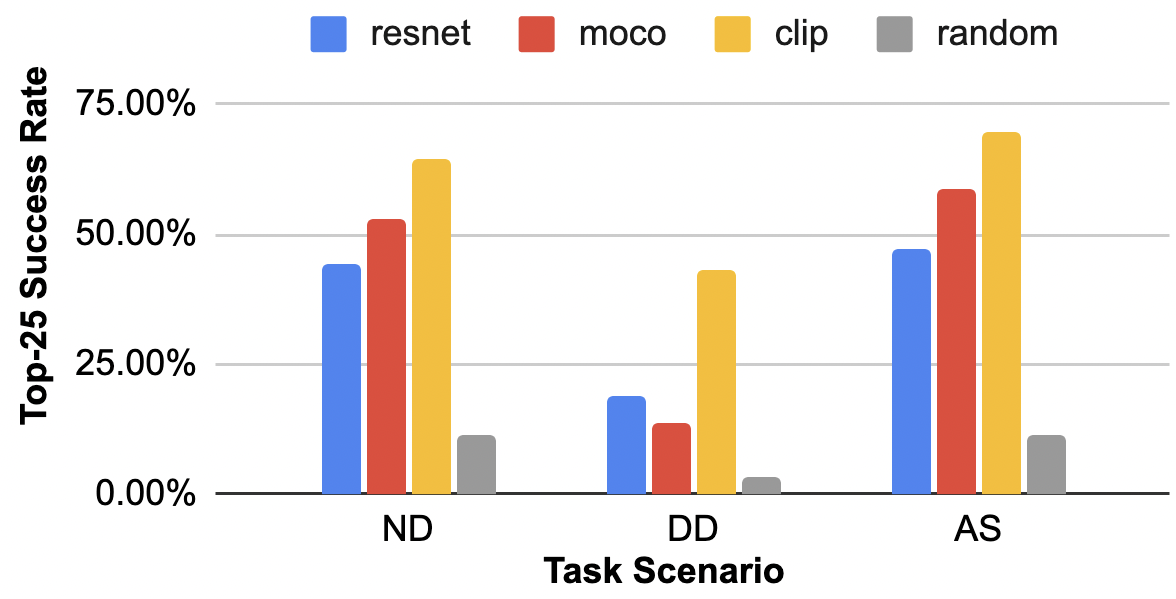}
    \vspace{-0.5cm}
    \caption{Top-25 success rate of embedding models with \textbf{Cos+Delta} under different task scenarios in Franka Kitchen with online images as goal specification modality. }
    \label{fig:franka_exps}
\end{figure}

The experimental results are presented in Figures~\ref{fig:embedding_model_dtv}~and~\ref{fig:franka_exps}. In terms of dataset total variation, we find that CLIP has the lowest total variation among the three embedding models across all the goal-specification modalities. We hypothesize that this is due to CLIP being trained on a larger data corpus (400 million images) compared to ResNet50 and MoCo trained on ImageNet (15 million images). CLIP has also shown promising zero-shot results in downstream tasks, and our observations in the goal selection tasks are consistent with these general trends. We also observe that using same scene images leads to better results than using internet images or drawings, which is along expected lines due to smaller domain gap. The Top-25 success rates for both the goal selection and high-level action selection tasks are plotted in Figure~\ref{fig:franka_exps}, along with the performance of a random selection mechanism, which can be used to judge the level of difficulty of the task. We find that ZeST with all the embedding models vastly outperform a random selection scheme suggesting that ZeST is capable of computing useful similarity scores. We again observe that CLIP is the best embedding model, resulting in a near 14-fold improvement over a random selection baseline in the harder DD setting.
Further details about this experiment can be found 
in the Appendix.

\subsection{How does ZeST perform in real-world video datasets?}
\label{subsec:realdata}

In this experiment, we examine whether our proposed approach can generalize to real world video data and how different embedding models compare.
% We hypothesize that there is major domain gap between still images and videos such that it is challenging to generalize ZeST with image-based foundation models to work on video dataset. 
To this end, we adopt a challenging real-world video dataset, SomethingSomething-V2~\citep{goyal2017something}, that consists of short clips of everyday human actions. These videos are recorded with hand-held devices and therefore do not have a stable view of the scene. Both the motion and human hand in the view can cause occlusion. 
We extract a subset of these video clips that all are \textit{opening something} as the dataset for experimental evaluation.
In addition to real video frames, we further test how synthetic images from a simulated environment interact with real world data, to test cross-domain task specification. Using objects from 3D Warehouse dataset, we generate object trajectories in simulation that matches with the \textit{opening something} tasks from the SomethingSomething-V2 dataset.
% We hypothesize that cross-domain task specification is particularly hard.

\begin{figure}
    \centering    
    \includegraphics[width=0.5\textwidth]{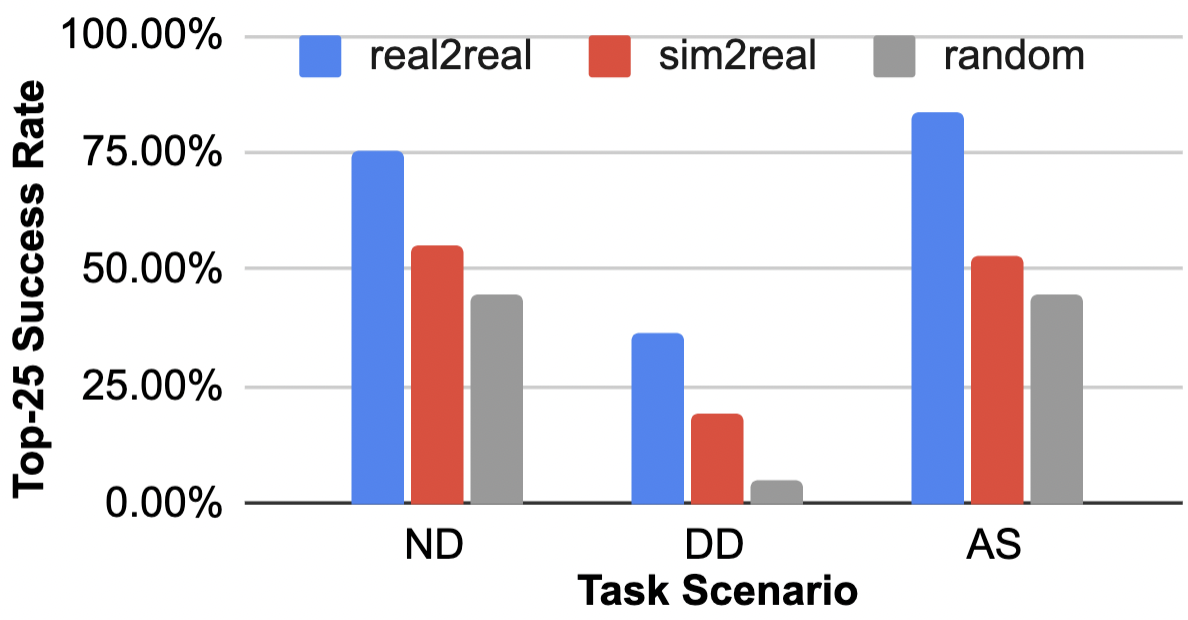}
    \vspace{-0.35cm}
    \caption{Top-25 success rate of \textbf{Cos+Delta} under different task scenarios with different task specification modalities in SSV2 dataset.}
    \label{fig:ssv2_tasks}
\end{figure}

\begin{figure}
    \centering
    \includegraphics[width=0.5\textwidth]{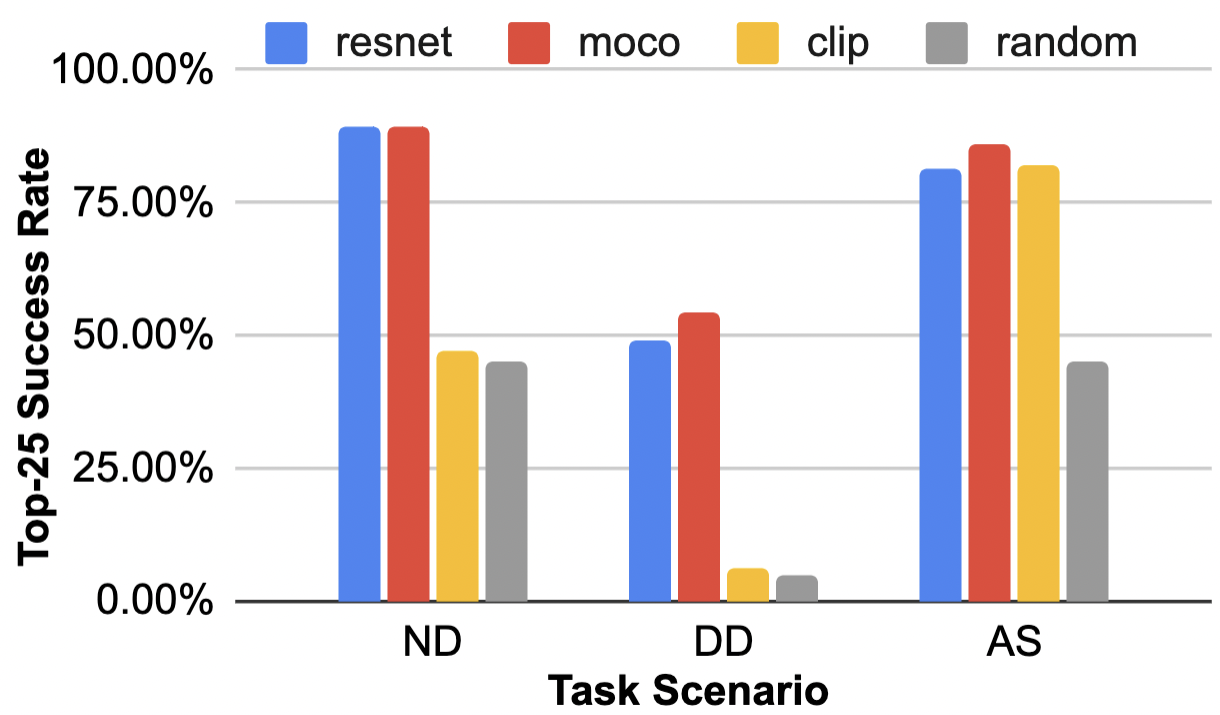}
    \vspace{-0.35cm}
    \caption{Top-25 success rate of different embedding models under different task scenarios in \textbf{real2real} setting in SSV2 dataset.}
    \label{fig:ssv2_exps}
\end{figure}

%We tested on both action selection and goal selection tasks and measured the top-10 success rates (note that goal selection is a much harder task). 
Similar to the evaluation in Franka kitchen domain, we consider narrow (ND) and diverse (DD) datasets for evaluation. In real2real modality of goal-specification, we use the frames from real trajectories to specify the goal while in the sim2real modality, we use rendered object frames to specify goal.
As shown in Figure~\ref{fig:ssv2_tasks}, averaging over all three embedding models, ZeST instantiations are able to outperform random guessing. Please see the 
Appendix for additional details and visualizations. Figure~\ref{fig:ssv2_exps} presents the results for individual embedding models in the real2real setting. Interestingly, we observe that CLIP is only marginally better than random guessing while ResNet and MoCo are substantially better. Overall, we observe that performing goal-selection and action-selection using real-world video datasets remains challenging. An interesting direction for future work would involve the learning of foundation models using internet-scale video datasets, which we hypothesize would lead to improved results.

\subsection{Can ZeST enable offline reinforcement learning?}

In this experiment, we test how ZeST signals could be used to enable policy learning in offline RL~\citep{Levine2020OfflineRL}. Offline RL is the setting where an agent is presented with a dataset of sub-optimal trajectories, and the agent must learn a competent policy using this dataset without additional environment interactions. We study if the similarity computed by ZeST can be used as a proxy for the reward function in offline RL. To do so, we use the Franka Kitchen domain outlined in Section~5.1, and use internet images for goal-specification. For our experimental evaluation, we compare: (1) behavior cloning (BC) which performs straightforward supervised learning to prediction actions in the dataset conditioned on observations; and (2) decision transformer (DT)~\citep{chen2021decisiontransformer} -- a state of the art offline RL algorithm that performs reward conditioned behavior cloning using a sequence model. This comparison involves two closely related algorithms with the major difference being reward conditioning. In the case of DT, we use our ZeST similarity as the reward function. Table~\ref{tab:rl_performance} shows the learned policies' average performance across task. When goal images with high similarity are selected, DT achieves 90.77\% of expert performance. This suggests that ZeST similarities serve as a reasonable proxy, and has the potential to enable offline RL.

\begin{table}[h!]
\centering
\begin{tabular}{l|ccc}
                  & \textbf{ZeST+DT} & \textbf{Demo.} & \textbf{BC}  \\ \hline
normalized return & 90.77\%     & 81.19\%   & 80.67\%
\end{tabular}
\caption{Normalized return for ZeST+DT comparing with demonstration dataset average and behavior cloning baseline (expert performance is 100\%).}
    \label{tab:rl_performance}
\end{table}

\pagebreak
\section{Conclusion}
\label{sec:conclusion}

In this work, we introduce a \textit{framework} for studying foundation models for zero-shot task specification (ZeST). 
We evaluate the effectiveness of ZeST for enabling zero-shot \textit{policy execution} through a set of goal selection tasks; 
and we evaluate ZeST for enabling \textit{policy learning} in offline reinforcement learning.
Leveraging existing pre-trained models, we demonstrate the effectiveness of ZeST on goal and action selection tasks in several different domains with different task specification modalities. 
We find that ZeST is quite effective in zero-shot goal-selection and results in a 14-fold increase in performance over a random guessing baseline. In offline RL, we find that using ZeST scores as a proxy for the reward function enables the learning of policies that perform better than a behavior cloning baseline.
Our results show the potential of foundation models and their applications in closing the domain gap for robot learning between real world and simulation.

Instantiations of ZeST in this work are limited in several aspects. The embedding model can be sensitive to occlusions, which often requires removing the robot arm from the camera view. Existing computer vision techniques for object segmentation and in-painting can be leveraged to address this problem for real applications. 
At the same time, our proposed approach is limited to tasks that have salient visual feature changes from unsuccessful states to successful states and such visual feature is not positional/location-based. However, explicit localization modules can be developed to address such issue and we leave this for future investigation. 

\newpage

\bibliography{conference}

\begin{thebibliography}{43}
\providecommand{\natexlab}[1]{#1}
\providecommand{\url}[1]{\texttt{#1}}
\expandafter\ifx\csname urlstyle\endcsname\relax
  \providecommand{\doi}[1]{doi: #1}\else
  \providecommand{\doi}{doi: \begingroup \urlstyle{rm}\Url}\fi

\bibitem[Amodei et~al.(2016)Amodei, Olah, Steinhardt, Christiano, Schulman, and
  Man{\'e}]{amodei2016concrete}
Dario Amodei, Chris Olah, Jacob Steinhardt, Paul Christiano, John Schulman, and
  Dan Man{\'e}.
\newblock Concrete problems in ai safety.
\newblock \emph{arXiv preprint arXiv:1606.06565}, 2016.

\bibitem[Andrychowicz et~al.(2017)Andrychowicz, Wolski, Ray, Schneider, Fong,
  Welinder, McGrew, Tobin, Abbeel, and Zaremba]{andrychowicz2017hindsight}
Marcin Andrychowicz, Filip Wolski, Alex Ray, Jonas Schneider, Rachel Fong,
  Peter Welinder, Bob McGrew, Josh Tobin, Pieter Abbeel, and Wojciech Zaremba.
\newblock Hindsight experience replay.
\newblock \emph{arXiv preprint arXiv:1707.01495}, 2017.

\bibitem[Ben-Shabat et~al.(2020)Ben-Shabat, Yu, Saleh, Campbell,
  Rodriguez-Opazo, Li, and Gould]{ikeadataset}
Yizhak Ben-Shabat, Xin Yu, Fatemehsadat Saleh, Dylan Campbell, Cristian
  Rodriguez-Opazo, Hongdong Li, and Stephen Gould.
\newblock The ikea asm dataset: Understanding people assembling furniture
  through actions, objects and pose.
\newblock 2020.

\bibitem[Bommasani et~al.(2021)Bommasani, Hudson, Adeli, Altman, Arora, von
  Arx, Bernstein, Bohg, Bosselut, Brunskill,
  et~al.]{bommasani2021opportunities}
Rishi Bommasani, Drew~A Hudson, Ehsan Adeli, Russ Altman, Simran Arora, Sydney
  von Arx, Michael~S Bernstein, Jeannette Bohg, Antoine Bosselut, Emma
  Brunskill, et~al.
\newblock On the opportunities and risks of foundation models.
\newblock \emph{arXiv preprint arXiv:2108.07258}, 2021.

\bibitem[Brown et~al.(2019)Brown, Goo, Nagarajan, and
  Niekum]{brown2019extrapolating}
Daniel Brown, Wonjoon Goo, Prabhat Nagarajan, and Scott Niekum.
\newblock Extrapolating beyond suboptimal demonstrations via inverse
  reinforcement learning from observations.
\newblock In \emph{International conference on machine learning}, pages
  783--792. PMLR, 2019.

\bibitem[Brown et~al.(2020)]{Brown2020GPT3}
Tom~B. Brown et~al.
\newblock Language models are few-shot learners.
\newblock \emph{ArXiv}, abs/2005.14165, 2020.

\bibitem[Bugliarello et~al.(2021)Bugliarello, Cotterell, Okazaki, and
  Elliott]{bugliarello2021multimodal}
Emanuele Bugliarello, Ryan Cotterell, Naoaki Okazaki, and Desmond Elliott.
\newblock Multimodal pretraining unmasked: A meta-analysis and a unified
  framework of vision-and-language berts.
\newblock \emph{Transactions of the Association for Computational Linguistics},
  9:\penalty0 978--994, 2021.

\bibitem[Chen et~al.(2021)Chen, Lu, Rajeswaran, Lee, Grover, Laskin, Abbeel,
  Srinivas, and Mordatch]{chen2021decisiontransformer}
Lili Chen, Kevin Lu, Aravind Rajeswaran, Kimin Lee, Aditya Grover, Michael
  Laskin, Pieter Abbeel, Aravind Srinivas, and Igor Mordatch.
\newblock Decision transformer: Reinforcement learning via sequence modeling.
\newblock \emph{arXiv preprint arXiv:2106.01345}, 2021.

\bibitem[Conde and Turgutlu(2021)]{conde2021clip}
Marcos~V Conde and Kerem Turgutlu.
\newblock Clip-art: Contrastive pre-training for fine-grained art
  classification.
\newblock In \emph{Proceedings of the IEEE/CVF Conference on Computer Vision
  and Pattern Recognition}, pages 3956--3960, 2021.

\bibitem[Cui(2021)]{cui2021dataset}
Yuchen Cui.
\newblock Visual task dataset.
\newblock \url{https://github.com/yuchen93/visual_tasks_dataset}, 2021.

\bibitem[Deng et~al.(2009)Deng, Dong, Socher, Li, Li, and
  Fei-Fei]{deng2009imagenet}
Jia Deng, Wei Dong, Richard Socher, Li-Jia Li, Kai Li, and Li~Fei-Fei.
\newblock Imagenet: A large-scale hierarchical image database.
\newblock In \emph{2009 IEEE conference on computer vision and pattern
  recognition}, pages 248--255. Ieee, 2009.

\bibitem[Devlin et~al.(2019)Devlin, Chang, Lee, and Toutanova]{Devlin2019BERT}
Jacob Devlin, Ming-Wei Chang, Kenton Lee, and Kristina Toutanova.
\newblock Bert: Pre-training of deep bidirectional transformers for language
  understanding.
\newblock In \emph{NAACL}, 2019.

\bibitem[Du et~al.(2019)Du, Krishnamurthy, Jiang, Agarwal, Dud{\'i}k, and
  Langford]{Du2019ProvablyER}
Simon~Shaolei Du, Akshay Krishnamurthy, Nan Jiang, Alekh Agarwal, Miroslav
  Dud{\'i}k, and John Langford.
\newblock Provably efficient rl with rich observations via latent state
  decoding.
\newblock \emph{ArXiv}, abs/1901.09018, 2019.

\bibitem[Eysenbach et~al.(2021)Eysenbach, Salakhutdinov, and
  Levine]{Eysenbach2021CLearningLT}
Benjamin Eysenbach, Ruslan Salakhutdinov, and Sergey Levine.
\newblock C-learning: Learning to achieve goals via recursive classification.
\newblock \emph{ArXiv}, abs/2011.08909, 2021.

\bibitem[Frans et~al.(2021)Frans, Soros, and Witkowski]{frans2021clipdraw}
Kevin Frans, LB~Soros, and Olaf Witkowski.
\newblock Clipdraw: Exploring text-to-drawing synthesis through language-image
  encoders.
\newblock \emph{arXiv preprint arXiv:2106.14843}, 2021.

\bibitem[Fu et~al.(2018)Fu, Singh, Ghosh, Yang, and Levine]{fu2018variational}
Justin Fu, Avi Singh, Dibya Ghosh, Larry Yang, and Sergey Levine.
\newblock Variational inverse control with events: A general framework for
  data-driven reward definition.
\newblock \emph{Advances in Neural Information Processing Systems},
  31:\penalty0 8538--8547, 2018.

\bibitem[Goyal et~al.(2017)Goyal, Ebrahimi~Kahou, Michalski, Materzynska,
  Westphal, Kim, Haenel, Fruend, Yianilos, Mueller-Freitag,
  et~al.]{goyal2017something}
Raghav Goyal, Samira Ebrahimi~Kahou, Vincent Michalski, Joanna Materzynska,
  Susanne Westphal, Heuna Kim, Valentin Haenel, Ingo Fruend, Peter Yianilos,
  Moritz Mueller-Freitag, et~al.
\newblock The" something something" video database for learning and evaluating
  visual common sense.
\newblock In \emph{Proceedings of the IEEE international conference on computer
  vision}, pages 5842--5850, 2017.

\bibitem[Gupta et~al.(2018)Gupta, Murali, Gandhi, and Pinto]{gupta2018robot}
Abhinav Gupta, Adithyavairavan Murali, Dhiraj~Prakashchand Gandhi, and Lerrel
  Pinto.
\newblock Robot learning in homes: Improving generalization and reducing
  dataset bias.
\newblock \emph{Advances in Neural Information Processing Systems},
  31:\penalty0 9094--9104, 2018.

\bibitem[Gupta et~al.(2019)Gupta, Kumar, Lynch, Levine, and
  Hausman]{gupta2019relay}
Abhishek Gupta, Vikash Kumar, Corey Lynch, Sergey Levine, and Karol Hausman.
\newblock Relay policy learning: Solving long-horizon tasks via imitation and
  reinforcement learning.
\newblock \emph{arXiv preprint arXiv:1910.11956}, 2019.

\bibitem[He et~al.(2016)He, Zhang, Ren, and Sun]{he2016deep}
Kaiming He, Xiangyu Zhang, Shaoqing Ren, and Jian Sun.
\newblock Deep residual learning for image recognition.
\newblock In \emph{Proceedings of the IEEE conference on computer vision and
  pattern recognition}, pages 770--778, 2016.

\bibitem[He et~al.(2020)He, Fan, Wu, Xie, and Girshick]{he2020momentum}
Kaiming He, Haoqi Fan, Yuxin Wu, Saining Xie, and Ross Girshick.
\newblock Momentum contrast for unsupervised visual representation learning.
\newblock In \emph{Proceedings of the IEEE/CVF Conference on Computer Vision
  and Pattern Recognition}, pages 9729--9738, 2020.

\bibitem[Izacard and Grave(2020)]{izacard2020leveraging}
Gautier Izacard and Edouard Grave.
\newblock Leveraging passage retrieval with generative models for open domain
  question answering.
\newblock \emph{arXiv preprint arXiv:2007.01282}, 2020.

\bibitem[Johns(2021)]{johns2021coarse}
Edward Johns.
\newblock Coarse-to-fine imitation learning: Robot manipulation from a single
  demonstration.
\newblock \emph{Proceedings of the International Conference on Robot and
  Automation}, 2021.

\bibitem[Kaelbling(1993)]{kaelbling1993learning}
Leslie~Pack Kaelbling.
\newblock Learning to achieve goals.
\newblock In \emph{IJCAI}, pages 1094--1099. Citeseer, 1993.

\bibitem[Khandelwal et~al.(2021)Khandelwal, Weihs, Mottaghi, and
  Kembhavi]{khandelwal2021simple}
Apoorv Khandelwal, Luca Weihs, Roozbeh Mottaghi, and Aniruddha Kembhavi.
\newblock Simple but effective: Clip embeddings for embodied ai.
\newblock \emph{arXiv preprint arXiv:2111.09888}, 2021.

\bibitem[Krizhevsky et~al.(2012)Krizhevsky, Sutskever, and
  Hinton]{Krizhevsky2012ImageNetCW}
Alex Krizhevsky, Ilya Sutskever, and Geoffrey~E. Hinton.
\newblock Imagenet classification with deep convolutional neural networks.
\newblock \emph{Communications of the ACM}, 60:\penalty0 84 -- 90, 2012.

\bibitem[Lee et~al.(2021)Lee, Hu, and Lim]{lee2021ikea}
Youngwoon Lee, Edward~S Hu, and Joseph~J Lim.
\newblock {IKEA} furniture assembly environment for long-horizon complex
  manipulation tasks.
\newblock In \emph{IEEE International Conference on Robotics and Automation
  (ICRA)}, 2021.
\newblock URL \url{https://clvrai.com/furniture}.

\bibitem[Levine et~al.(2020)Levine, Kumar, Tucker, and Fu]{Levine2020OfflineRL}
Sergey Levine, Aviral Kumar, G.~Tucker, and Justin Fu.
\newblock Offline reinforcement learning: Tutorial, review, and perspectives on
  open problems.
\newblock \emph{ArXiv}, abs/2005.01643, 2020.

\bibitem[Lin et~al.(2021)Lin, Nogueira, and Yates]{lin2021pretrained}
Jimmy Lin, Rodrigo Nogueira, and Andrew Yates.
\newblock Pretrained transformers for text ranking: Bert and beyond.
\newblock \emph{Synthesis Lectures on Human Language Technologies}, 14\penalty0
  (4):\penalty0 1--325, 2021.

\bibitem[Mandlekar et~al.(2020)Mandlekar, Xu, Mart{\i}n-Mart{\i}n, Savarese,
  and Fei-Fei]{mandlekargti}
Ajay Mandlekar, Danfei Xu, Roberto Mart{\i}n-Mart{\i}n, Silvio Savarese, and
  Li~Fei-Fei.
\newblock Gti: Learning to generalize across long-horizon tasks from human
  demonstrations.
\newblock \emph{Proceedings of Robotics: Science and Systems}, 2020.

\bibitem[Nair et~al.(2018)Nair, Pong, Dalal, Bahl, Lin, and
  Levine]{nair2018visual}
Ashvin~V Nair, Vitchyr Pong, Murtaza Dalal, Shikhar Bahl, Steven Lin, and
  Sergey Levine.
\newblock Visual reinforcement learning with imagined goals.
\newblock \emph{Advances in Neural Information Processing Systems},
  31:\penalty0 9191--9200, 2018.

\bibitem[Nair and Finn(2019)]{nair2019hierarchical}
Suraj Nair and Chelsea Finn.
\newblock Hierarchical foresight: Self-supervised learning of long-horizon
  tasks via visual subgoal generation.
\newblock In \emph{International Conference on Learning Representations}, 2019.

\bibitem[Parisi et~al.(2022)Parisi, Rajeswaran, Purushwalkam, and
  Gupta]{Parisi2022PVR}
Simone Parisi, Aravind Rajeswaran, Senthil Purushwalkam, and Abhinav Gupta.
\newblock {The Unsurprising Effectiveness of Pre-Trained Vision Models for
  Control}.
\newblock \emph{ArXiv}, abs/2203.03580, 2022.

\bibitem[Patashnik et~al.(2021)Patashnik, Wu, Shechtman, Cohen-Or, and
  Lischinski]{patashnik2021styleclip}
Or~Patashnik, Zongze Wu, Eli Shechtman, Daniel Cohen-Or, and Dani Lischinski.
\newblock Styleclip: Text-driven manipulation of stylegan imagery.
\newblock In \emph{Proceedings of the IEEE/CVF International Conference on
  Computer Vision}, pages 2085--2094, 2021.

\bibitem[Pinto and Gupta(2016)]{Pinto2016SupersizingSL}
Lerrel Pinto and Abhinav~Kumar Gupta.
\newblock Supersizing self-supervision: Learning to grasp from 50k tries and
  700 robot hours.
\newblock \emph{2016 IEEE International Conference on Robotics and Automation
  (ICRA)}, pages 3406--3413, 2016.

\bibitem[Radford et~al.(2019)Radford, Wu, Child, Luan, Amodei, and
  Sutskever]{Radford2019GPT}
Alec Radford, Jeff Wu, Rewon Child, David Luan, Dario Amodei, and Ilya
  Sutskever.
\newblock Language models are unsupervised multitask learners.
\newblock 2019.

\bibitem[Radford et~al.(2021)Radford, Kim, Hallacy, Ramesh, Goh, Agarwal,
  Sastry, Askell, Mishkin, Clark, et~al.]{radford2021learning}
Alec Radford, Jong~Wook Kim, Chris Hallacy, Aditya Ramesh, Gabriel Goh,
  Sandhini Agarwal, Girish Sastry, Amanda Askell, Pamela Mishkin, Jack Clark,
  et~al.
\newblock Learning transferable visual models from natural language
  supervision.
\newblock \emph{arXiv preprint arXiv:2103.00020}, 2021.

\bibitem[Rajeswaran et~al.(2017)Rajeswaran, Lowrey, Todorov, and
  Kakade]{rajeswaran17towards}
Aravind Rajeswaran, Kendall Lowrey, Emanuel Todorov, and Sham Kakade.
\newblock {Towards Generalization and Simplicity in Continuous Control}.
\newblock In \emph{NIPS}, 2017.

\bibitem[Rajeswaran et~al.(2018)Rajeswaran, Kumar, Gupta, Vezzani, Schulman,
  Todorov, and Levine]{rajeswaran18learning}
Aravind Rajeswaran, Vikash Kumar, Abhishek Gupta, Giulia Vezzani, John
  Schulman, Emanuel Todorov, and Sergey Levine.
\newblock {Learning Complex Dexterous Manipulation with Deep Reinforcement
  Learning and Demonstrations}.
\newblock In \emph{Proceedings of Robotics: Science and Systems (RSS)}, 2018.

\bibitem[Shah and Kumar(2021)]{shah2021rrl}
Rutav Shah and Vikash Kumar.
\newblock Rrl: Resnet as representation for reinforcement learning.
\newblock In \emph{Self-Supervision for Reinforcement Learning Workshop-ICLR
  2021}, 2021.

\bibitem[Shridhar et~al.(2021)Shridhar, Manuelli, and Fox]{shridhar2021cliport}
Mohit Shridhar, Lucas Manuelli, and Dieter Fox.
\newblock Cliport: What and where pathways for robotic manipulation.
\newblock In \emph{Proceedings of the 5th Conference on Robot Learning (CoRL)},
  2021.

\bibitem[Singh et~al.(2019)Singh, Yang, Hartikainen, Finn, and
  Levine]{singh2019end}
Avi Singh, Larry Yang, Kristian Hartikainen, Chelsea Finn, and Sergey Levine.
\newblock End-to-end robotic reinforcement learning without reward engineering.
\newblock \emph{Proceedings of Robotics: Science and Systems}, 2019.

\bibitem[Tian et~al.(2020)Tian, Wang, Krishnan, Tenenbaum, and
  Isola]{tian2020rethinking}
Yonglong Tian, Yue Wang, Dilip Krishnan, Joshua~B Tenenbaum, and Phillip Isola.
\newblock Rethinking few-shot image classification: a good embedding is all you
  need?
\newblock In \emph{Computer Vision--ECCV 2020: 16th European Conference,
  Glasgow, UK, August 23--28, 2020, Proceedings, Part XIV 16}, pages 266--282.
  Springer, 2020.

\end{thebibliography}

\newpage
\appendix

 % Add the appendix text to the document TOC
\addcontentsline{toc}{section}{Appendix}

\part{Appendix}
{\let\clearpage\relax \parttoc} % Start the appendix part
 % Insert the appendix TOC

\newpage
\section{Details of Foundation Models under Test}

%%AR.11.16 : We should formally cover the CLIP loss as well as MoCo in this section. We can then simply say that such self-supervision models have found wide applicability in CV as a backbone representations. However, they have not been widely used in RL, and our work explores this possibility.

Vision-based robot learning enables training end-to-end robot policies that directly work with visual inputs. However, real-world robot manipulation data are expensive and often cannot generate data at the scale required to train vision models. Therefore, existing vision-based robot learning methods often rely on pre-trained vision models and fine-tune with limited data.
Deep residual network (ResNet), proposed by \cite{he2016deep}, greatly increased the depth of neural networks through using residual connections.
ResNet has been used as the backbone for many state-of-the-art vision models. Pre-trained ResNet on image classification task on ImageNet~\citep{deng2009imagenet} has been a popular and very effective embedding model for extracting visual features for down-stream robotic manipulation tasks~\citep{gupta2018robot,mandlekargti,shah2021rrl}.
MoCo~\citep{he2020momentum} is a recent self-supervised pre-training technique, i.e. pre-train without using the labels, and has shown better performance for downstream vision tasks than supervised pre-training with image classification. 
Moco employs a contrastive loss, for some query $q$ and a set of encoded samples $\{k_0,k_1,...k_K\}$ (where $k_+$ is the desired key for query $q$):
\begin{equation}
   \mathcal{L}_q = - \log \frac{\exp{(q\cdot k_+/\tau)}}{\sum^K_{i=0}\exp(q\cdot k_i/\tau)}
\end{equation}
CLIP~\citep{radford2021learning} is a multi-modal pre-training technique that leverages both images and text descriptions to learn embeddings of both modality in the same space. 
The training objective of CLIP is to maximize the cosine similarity between matching pairs of encoded image $i_k$ and text $t_k$ while minimizing the similarities between unmatched pairs:
\begin{equation}
    \mathcal{L}_k = - \log \frac{\exp{(i_k\cdot t_k/\tau)}}{\sum^N_{n=0}\exp(i_k \cdot t_n/\tau)}
                    - \log \frac{\exp{(i_k\cdot t_k/\tau)}}{\sum^N_{n=0}\exp(i_n \cdot t_k/\tau)}
\end{equation}
CLIP is trained with over 400 million pairs of internet image and text and has shown state-of-the-art zero-shot performance on image classification tasks. 
Self-supervision models such as Moco and CLIP have found wide applicability in computer vision as backbone representations. However, they have not been widely used in RL, and our work explores this possibility.

\section{Further Analysis of ZeST and Additional Design Considerations} 

\begin{figure}
    \centering
    \includegraphics[width=0.85\textwidth]{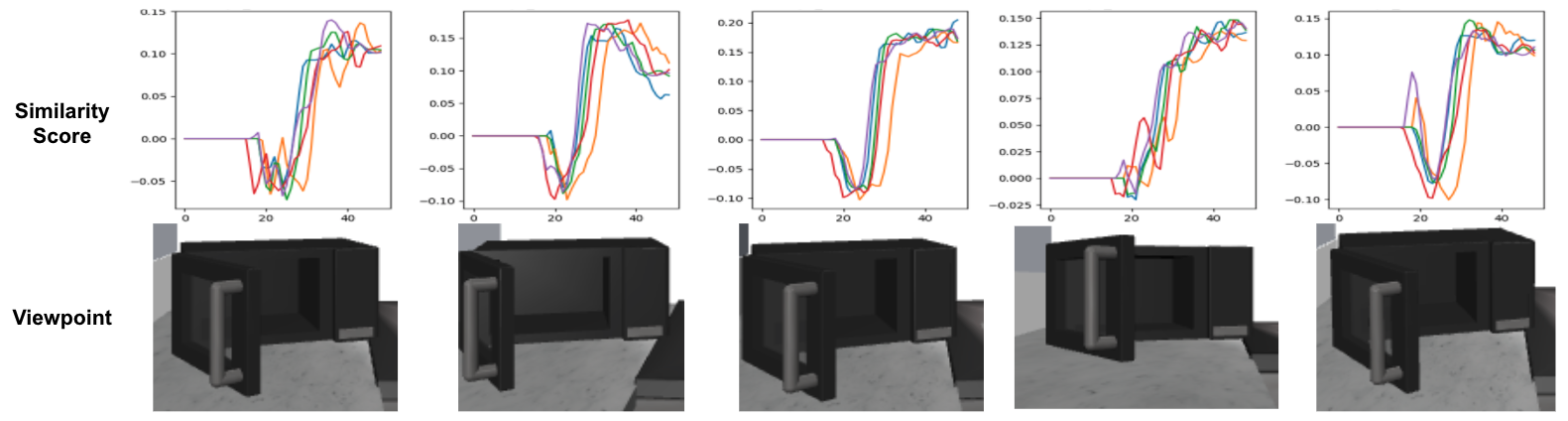}
    \caption{\textbf{Cos+Delta} signals under different viewpoints for microwave opening task.}
    \label{fig:viewpoints}
\end{figure}

\paragraph{Viewpoints}
To mitigate potential viewpoint differences between the observations and goal specification images, it is desired to use multi-viewpoint observations and take the average similarity score across different viewpoints.
As shown in Figure~\ref{fig:viewpoints}, 
we observe that the cosine similarity for same trajectories of opening a microwave differ from different viewpoints. However, this may not be possible in certain real world applications and can be mitigated by filtering or reweighing goal specifications using  similarities with embeddings of the robot's observations.

\begin{figure}
    \centering
    \includegraphics[width=0.75\textwidth]{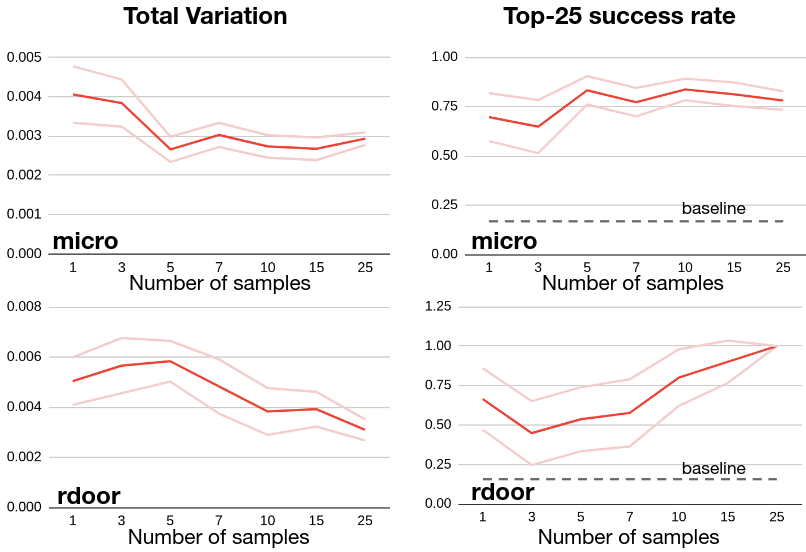}
    \caption{Example performance profiles based on number of goal samples used.}
    \label{fig:goal_samples}
\end{figure}

\paragraph{Ensemble of task specifications} To reduce potential noise in the computed signals, it is desired to use an \textbf{ensemble} of goal specification images. When multiple goal and initial images are provided, for each goal image, we will find an initial image that is the most similar to the goal image's embedding using cosine similarity measure and then compute the delta feature between the two. 
With this ensemble of images, the average of all cosine similarity scores is used as the final prediction. 
 
We find the use of an ensemble of goal specifications help to reduce noise and improve the smoothness of the predicted signal. 
We further investigate how the performance of one instantiation of ZeSTS change based on the number of goal samples used in the dataset. 
We evaluated with goal selection task in the Franka Kitchen domain using internet images as the goal specification modality. Each task has a total of 50 goal samples (for both the initial and goal state). We sample a desired number of goal samples from the dataset and repeat for 5 times. The average performance with standard errors is plotted in Figure~\ref{fig:goal_samples}. In general, more samples would improve the performance of the model, but the initial number of samples necessary to achieve a certain performance varies from task to task.

\paragraph{Pixel differences and \textbf{Cos+Delta} scores}

\begin{figure}
    \centering
    \includegraphics[width=0.9\textwidth]{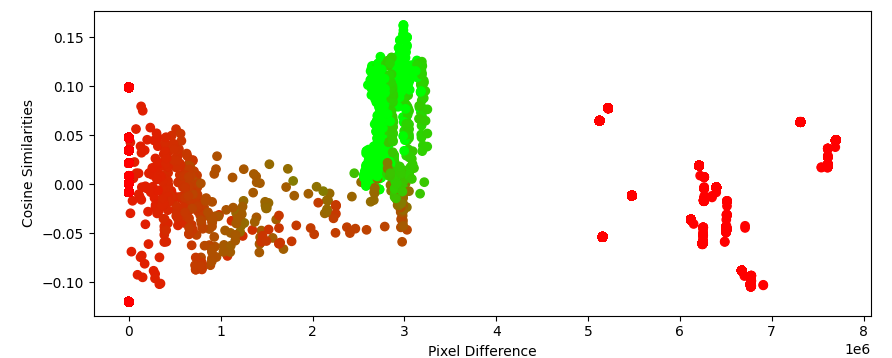}
    \caption{Relationship between cosine similarities between delta features and pixel differences between frames for microwave opening task in Franka Kitchen with CLIP embeddings (the greenness of a dot is indicative of its distance to true goal states).}
    \label{fig:pixel_diffs}
\end{figure}

A natural question to ask when considering whether cosine similarities between delta features are meaningful is that ``Does pixel differences between image frames correlate with high cosine similarities?". Figure~\ref{fig:pixel_diffs} shows that large pixel differences do not correlate with high similarity scores.

\section{Qualitative Evaluations of ZeST on Real Image Trajectories}

\begin{figure}
    \centering
    \includegraphics[width=\textwidth]{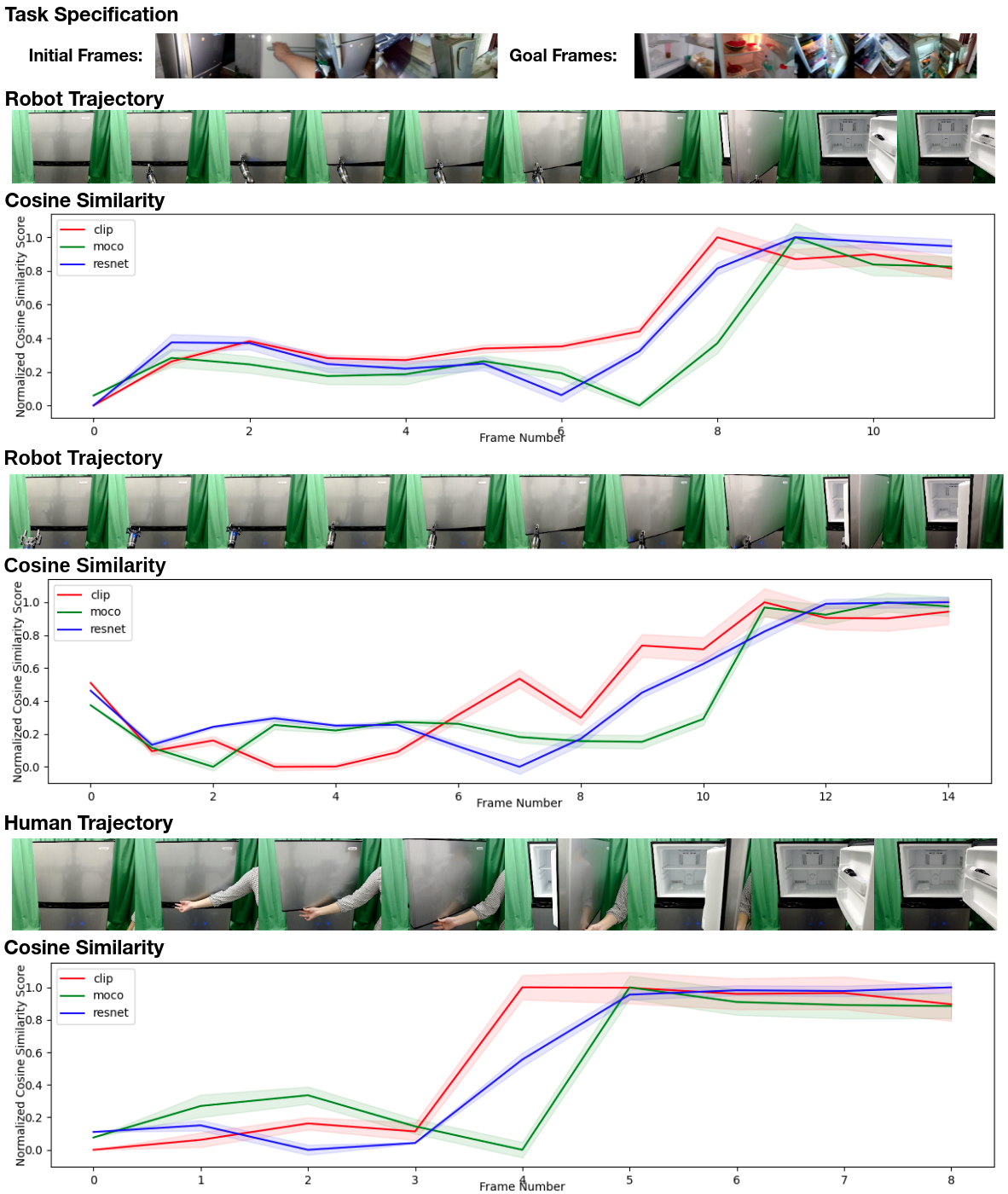}
    \caption{Opening Refrigerator Task (top door)}
    \label{fig:fridge1}
\end{figure}

\begin{figure}
    \centering
    \includegraphics[width=\textwidth]{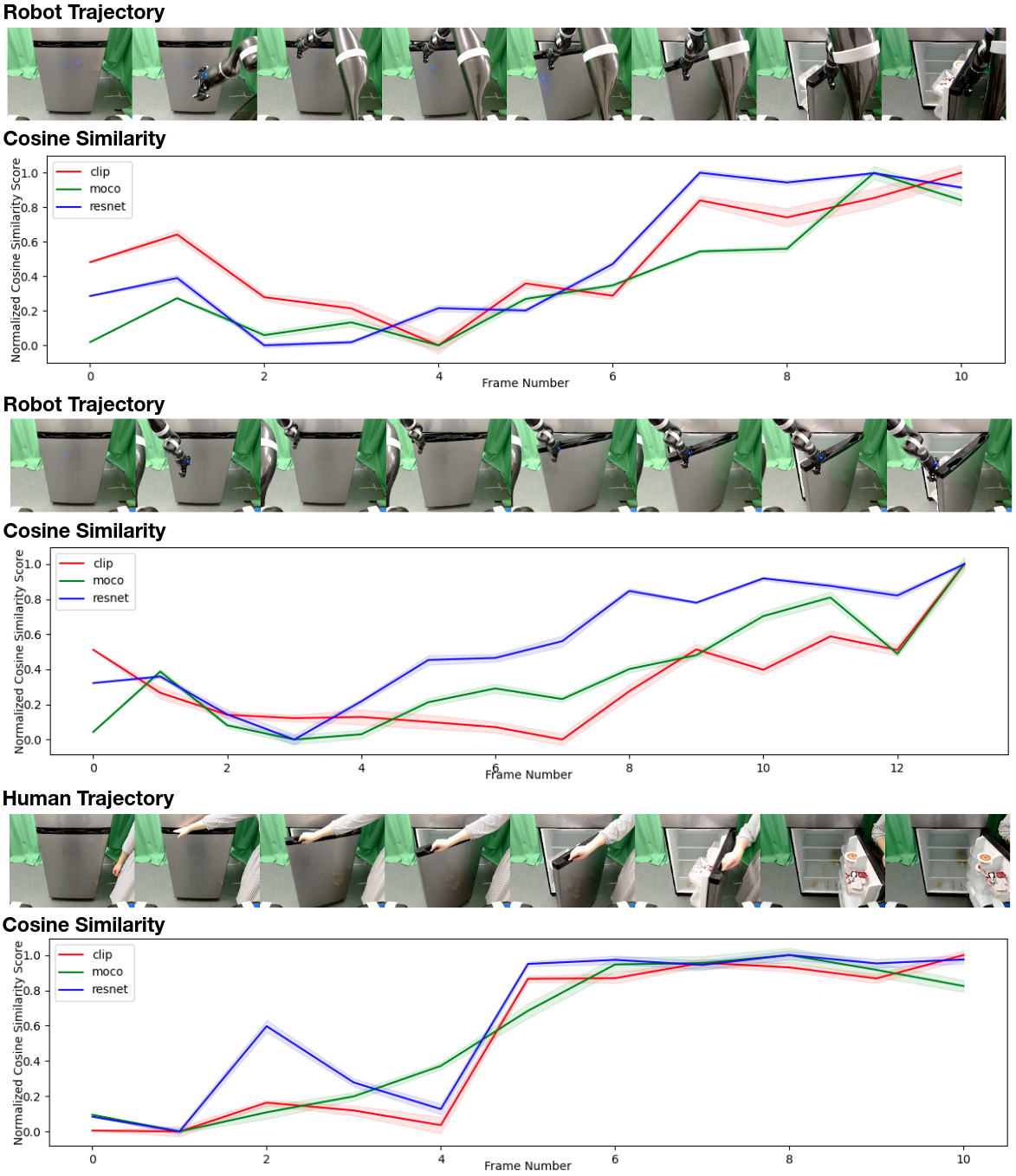}
    \caption{Opening Refrigerator Task (bottom door)}
    \label{fig:fridge2}
\end{figure}

\begin{figure}
    \centering
    \includegraphics[width=\textwidth]{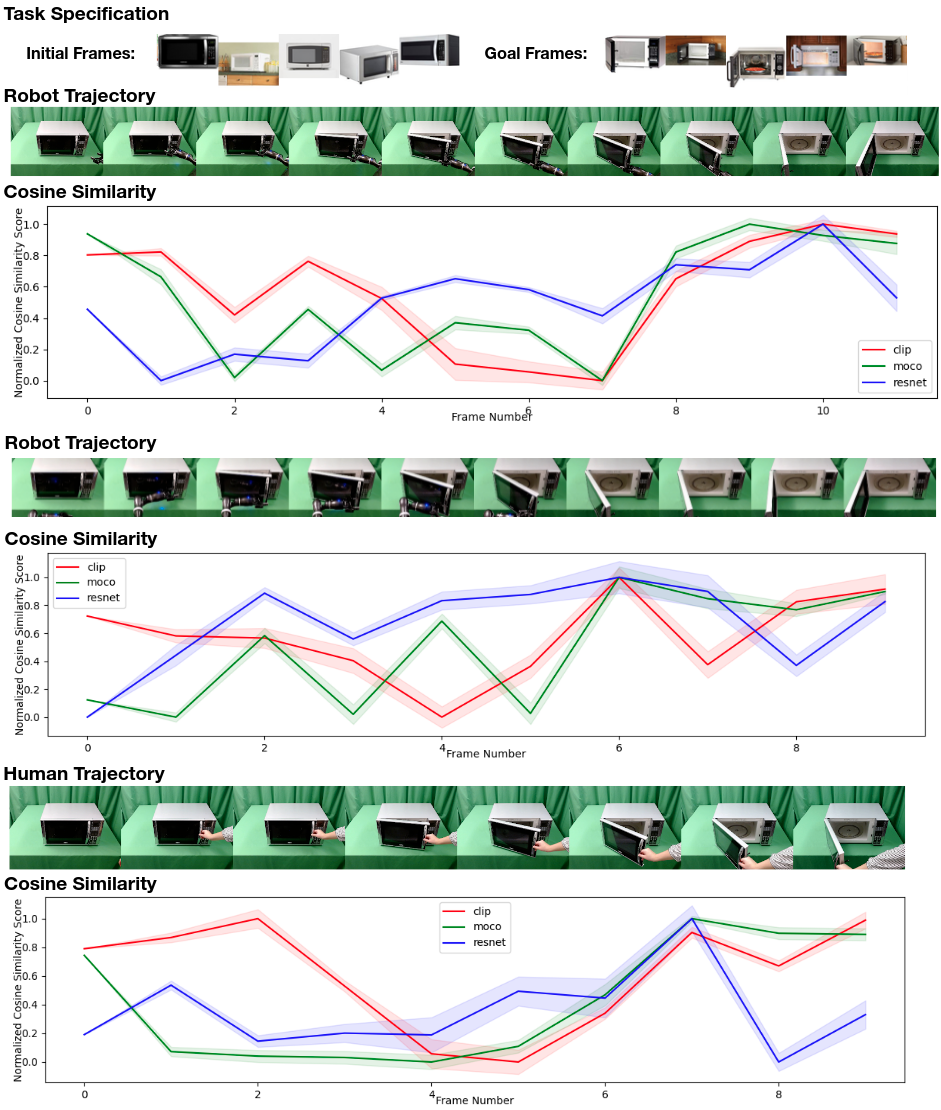}
    \caption{Opening Microwave Task}
    \label{fig:micro}
\end{figure}

\begin{figure}
    \centering
    \includegraphics[width=\textwidth]{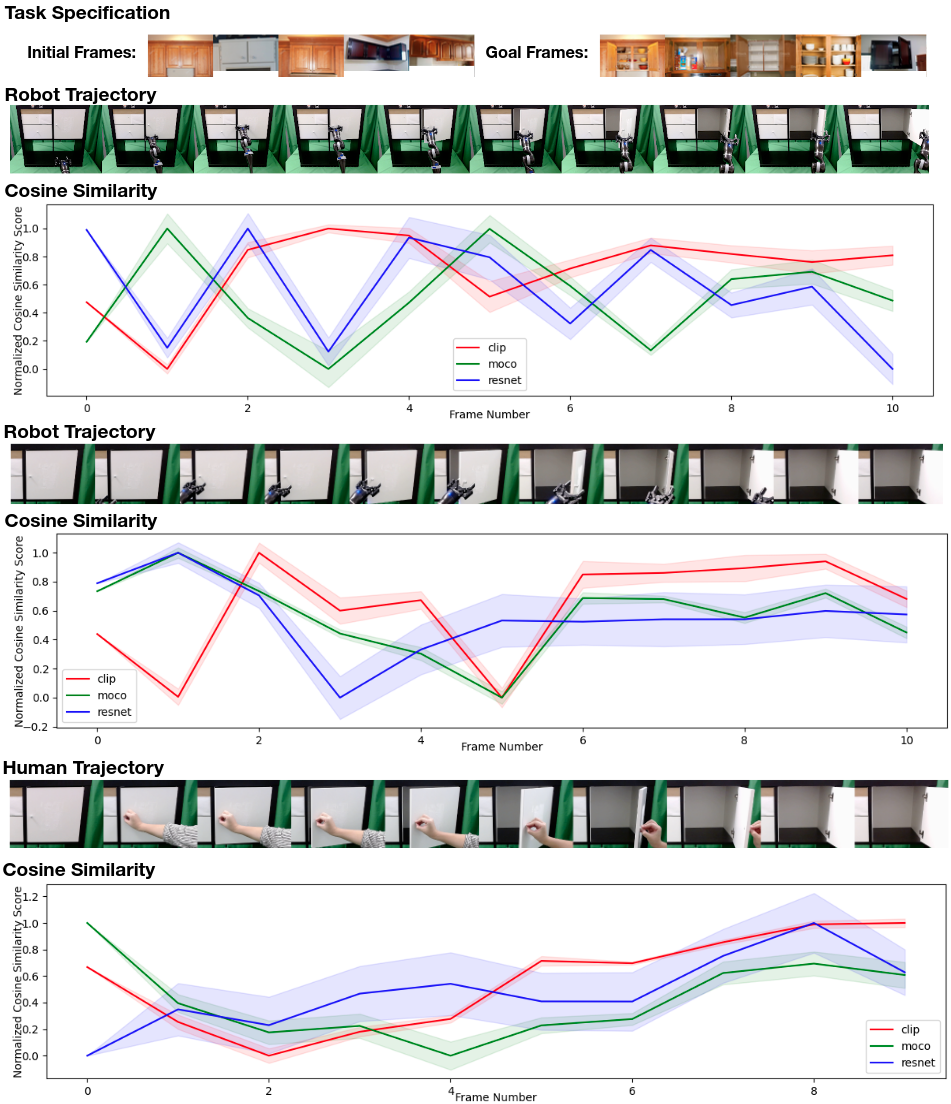}
    \caption{Opening Cabinet Task}
    \label{fig:cabinet}
\end{figure}

\begin{figure}
    \centering
    \includegraphics[width=\textwidth]{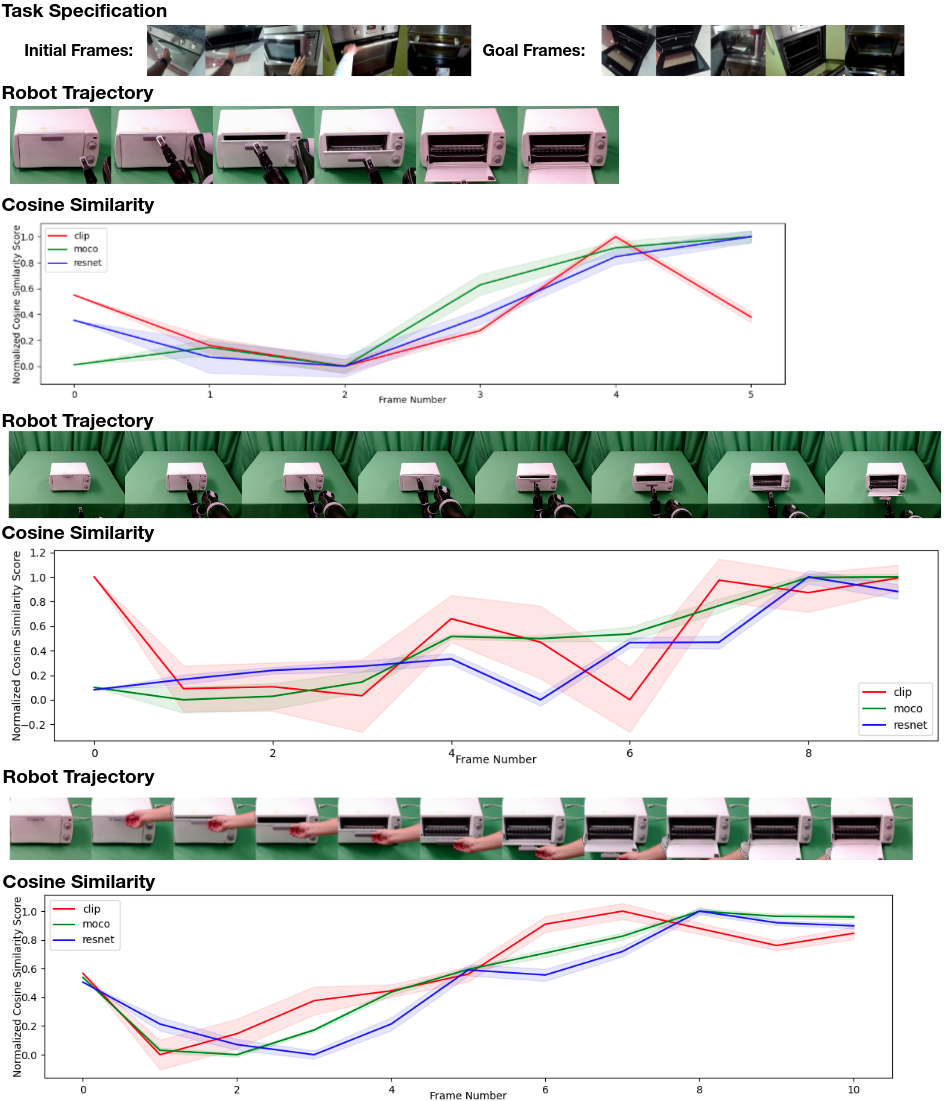}
    \caption{Opening Oven Task}
    \label{fig:oven}
\end{figure}

\begin{figure}
    \centering
    \includegraphics[width=\textwidth]{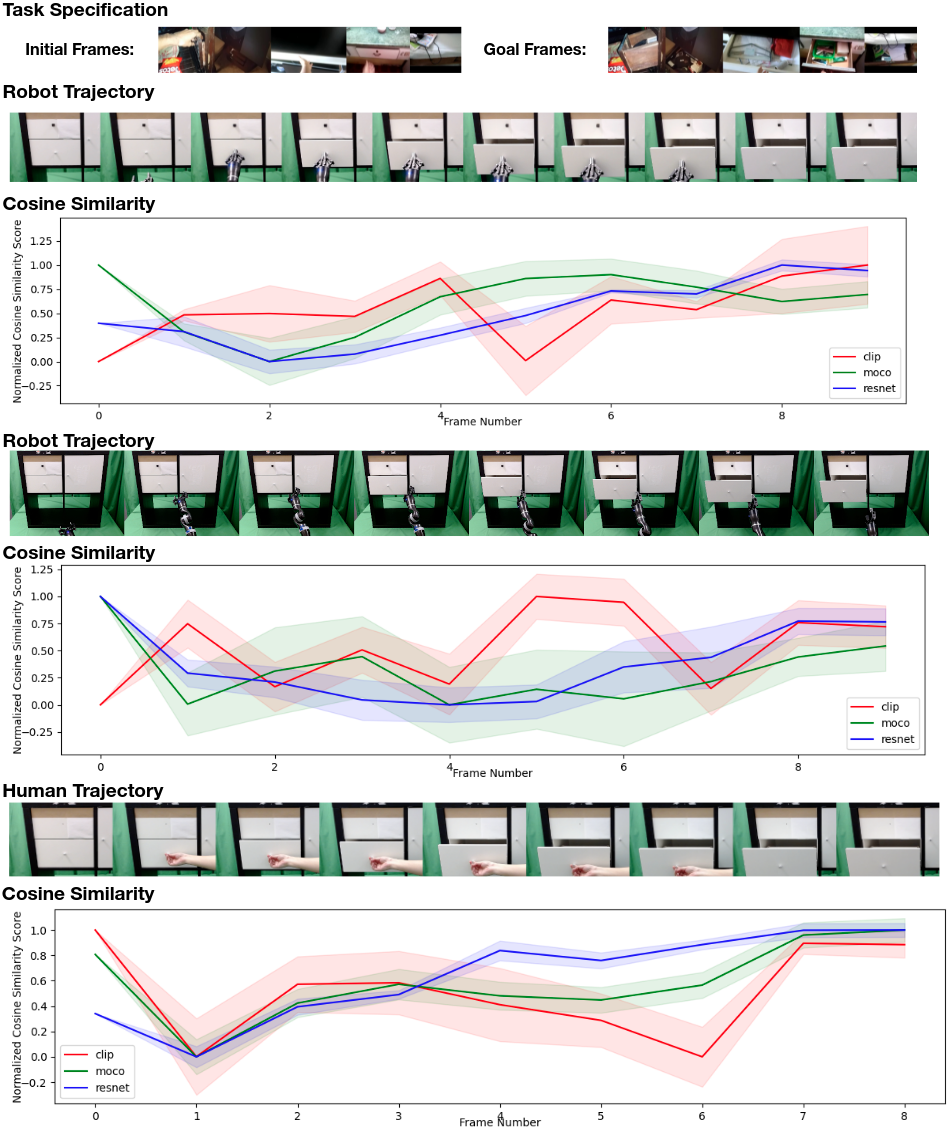}
    \caption{Opening Drawer Task}
    \label{fig:drawer}
\end{figure}

In this section we present qualitative evaluation of \textbf{Cos+Delta} with all three embedding models for real world object manipulation trajectories. Sample task specification images used, robot trajectory frames, and the corresponding normalized cosine similarity score are plotted. Figure~\ref{fig:micro},\ref{fig:cabinet},\ref{fig:fridge1},\ref{fig:fridge2},\ref{fig:oven}, and \ref{fig:drawer} shows the evaluation for tasks of opening fridge top door, opening fridge bottom door, opening a microwave, opening a cabinet,  opening an oven, and opening a drawer respectively. As shown in these figures, ZeST signals are sensitive to occlusions caused by the robot manipulator as well as different croppings of the trajectory images.

{\renewcommand{\arraystretch}{1.2}
\begin{table}[]
\centering
\scriptsize
\begin{tabular}{|cc|cccccc|c|c|}
\hline
\multicolumn{1}{|c|}{\multirow{2}{*}{\textbf{Modality}}} & \multirow{2}{*}{\textbf{model}} & \multicolumn{6}{c|}{\textbf{Tasks}}                                                                                                                                                                          & \multirow{2}{*}{\textbf{Avg.}} & \multirow{2}{*}{\textbf{Std. Err}} \\ \cline{3-8}
\multicolumn{1}{|c|}{}                                   &                                 & \multicolumn{1}{c|}{\textbf{knob3}} & \multicolumn{1}{c|}{\textbf{rdoor}} & \multicolumn{1}{c|}{\textbf{sdoor}} & \multicolumn{1}{c|}{\textbf{knob2}} & \multicolumn{1}{c|}{\textbf{ldoor}} & \textbf{micro} &                                &                                         \\ \hline
\multicolumn{2}{|c|}{Random (goal frame ratio)}                                            & \multicolumn{1}{c|}{10.90\%}        & \multicolumn{1}{c|}{5.3\%}          & \multicolumn{1}{c|}{33.4\%}         & \multicolumn{1}{c|}{4.9\%}          & \multicolumn{1}{c|}{5.7\%}          & 8.0\%          & 11.4\%                         & -                                       \\ \hline
\multicolumn{1}{|c|}{\multirow{3}{*}{same\_scene\_imgs}} & resnet                          & \multicolumn{1}{c|}{3.0\%}          & \multicolumn{1}{c|}{100.0\%}        & \multicolumn{1}{c|}{76.0\%}         & \multicolumn{1}{c|}{97.0\%}         & \multicolumn{1}{c|}{100.0\%}        & 90.0\%         & 77.7\%                         & 2.33\%                                  \\ \cline{2-10} 
\multicolumn{1}{|c|}{}                                   & moco                            & \multicolumn{1}{c|}{31.0\%}         & \multicolumn{1}{c|}{100.0\%}        & \multicolumn{1}{c|}{98.0\%}         & \multicolumn{1}{c|}{99.0\%}         & \multicolumn{1}{c|}{93.0\%}         & 100.0\%        & 86.8\%                         & 3.17\%                                  \\ \cline{2-10} 
\multicolumn{1}{|c|}{}                                   & clip                            & \multicolumn{1}{c|}{98.0\%}         & \multicolumn{1}{c|}{100.0\%}        & \multicolumn{1}{c|}{62.0\%}         & \multicolumn{1}{c|}{100.0\%}        & \multicolumn{1}{c|}{91.0\%}         & 94.0\%         & \textbf{90.8\%}                & 4.73\%                                  \\ \hline
\multicolumn{1}{|c|}{\multirow{3}{*}{online\_imgs}}      & resnet                          & \multicolumn{1}{c|}{0.0\%}          & \multicolumn{1}{c|}{28.3\%}         & \multicolumn{1}{c|}{36.0\%}         & \multicolumn{1}{c|}{34.3\%}         & \multicolumn{1}{c|}{21.0\%}         & 59.7\%         & 29.9\%                         & 14.44\%                                 \\ \cline{2-10} 
\multicolumn{1}{|c|}{}                                   & moco                            & \multicolumn{1}{c|}{26.3\%}         & \multicolumn{1}{c|}{79.3\%}         & \multicolumn{1}{c|}{1.0\%}          & \multicolumn{1}{c|}{67.7\%}         & \multicolumn{1}{c|}{8.3\%}          & 77.0\%         & 43.3\%                         & 11.35\%                                 \\ \cline{2-10} 
\multicolumn{1}{|c|}{}                                   & clip                            & \multicolumn{1}{c|}{17.7\%}         & \multicolumn{1}{c|}{83.7\%}         & \multicolumn{1}{c|}{80.3\%}         & \multicolumn{1}{c|}{13.3\%}         & \multicolumn{1}{c|}{77.0\%}         & 92.0\%         & \textbf{60.7\%}                & 8.95\%                                  \\ \hline
\multicolumn{1}{|c|}{\multirow{3}{*}{drawings}}          & resnet                          & \multicolumn{1}{c|}{100.0\%}        & \multicolumn{1}{c|}{0.0\%}          & \multicolumn{1}{c|}{0.0\%}          & \multicolumn{1}{c|}{33.0\%}         & \multicolumn{1}{c|}{0.0\%}          & 16.0\%         & 24.8\%                         & 2.95\%                                  \\ \cline{2-10} 
\multicolumn{1}{|c|}{}                                   & moco                            & \multicolumn{1}{c|}{27.0\%}         & \multicolumn{1}{c|}{98.0\%}         & \multicolumn{1}{c|}{12.0\%}         & \multicolumn{1}{c|}{12.5\%}         & \multicolumn{1}{c|}{0.0\%}          & 26.0\%         & 29.3\%                         & 4.92\%                                  \\ \cline{2-10} 
\multicolumn{1}{|c|}{}                                   & clip                            & \multicolumn{1}{c|}{24.5\%}         & \multicolumn{1}{c|}{0.0\%}          & \multicolumn{1}{c|}{49.0\%}         & \multicolumn{1}{c|}{23.0\%}         & \multicolumn{1}{c|}{98.0\%}         & 57.0\%         & \textbf{41.9\%}                & 10.33\%                                 \\ \hline
\multicolumn{1}{|c|}{text}                               & clip                            & \multicolumn{1}{c|}{0.0\%}          & \multicolumn{1}{c|}{46.0\%}         & \multicolumn{1}{c|}{4.0\%}          & \multicolumn{1}{c|}{0.0\%}          & \multicolumn{1}{c|}{8.0\%}          & 1.0\%          & 9.8\%                          & 2.30\%                                  \\ \hline
\end{tabular}
\caption{ Top-25 goal selection success rate in Franka kitchen domain (SGSD).}
     \label{tab:franka_kitchen}
     \vspace{-0.3cm}
\end{table}

}

{\renewcommand{\arraystretch}{1.2}

\begin{table}[]
    \centering
    \scriptsize
\begin{tabular}{|c|c|rrrrrr|r|}
\hline
\textbf{Task Specification} & \multirow{2}{*}{\textbf{model}} & \multicolumn{6}{c|}{\textbf{Task ID}}                                                                                                                                                                                             & \multicolumn{1}{c|}{\multirow{2}{*}{\textbf{avg}}} \\ \cline{3-8}
                   \textbf{modality}                                   &                                           & \multicolumn{1}{c|}{\textbf{knob3}} & \multicolumn{1}{c|}{\textbf{rdoor}} & \multicolumn{1}{c|}{\textbf{sdoor}} & \multicolumn{1}{c|}{\textbf{knob2}} & \multicolumn{1}{c|}{\textbf{ldoor}} & \multicolumn{1}{c|}{\textbf{micro}} & \multicolumn{1}{c|}{}                              \\ \hline
\multirow{3}{*}{same\_scene\_imgs}                    & ResNet                                    & \multicolumn{1}{r|}{26.40\%}        & \multicolumn{1}{r|}{23.20\%}        & \multicolumn{1}{r|}{53.60\%}        & \multicolumn{1}{r|}{40.80\%}        & \multicolumn{1}{r|}{20.00\%}        & 30.67\%                             & \textbf{32.44\%}                                   \\ \cline{2-9} 
                                                      & Moco                                      & \multicolumn{1}{r|}{27.20\%}        & \multicolumn{1}{r|}{19.20\%}        & \multicolumn{1}{r|}{3.20\%}         & \multicolumn{1}{r|}{39.20\%}        & \multicolumn{1}{r|}{35.33\%}        & 40.00\%                             & 27.36\%                                            \\ \cline{2-9} 
                                                      & CLIP                                      & \multicolumn{1}{r|}{18.67\%}        & \multicolumn{1}{r|}{6.40\%}         & \multicolumn{1}{r|}{7.20\%}         & \multicolumn{1}{r|}{28.00\%}        & \multicolumn{1}{r|}{28.80\%}        & 16.00\%                             & 17.51\%                                            \\ \hline
\multirow{3}{*}{online\_imgs}                         & ResNet                                    & \multicolumn{1}{r|}{23.47\%}        & \multicolumn{1}{r|}{5.07\%}         & \multicolumn{1}{r|}{20.00\%}        & \multicolumn{1}{r|}{3.73\%}         & \multicolumn{1}{r|}{11.20\%}        & 26.44\%                             & 14.99\%                                            \\ \cline{2-9} 
                                                      & Moco                                      & \multicolumn{1}{r|}{26.40\%}        & \multicolumn{1}{r|}{20.00\%}        & \multicolumn{1}{r|}{3.20\%}         & \multicolumn{1}{r|}{41.33\%}        & \multicolumn{1}{r|}{31.73\%}        & 39.78\%                             & \textbf{27.07\%}                                   \\ \cline{2-9} 
                                                      & CLIP                                      & \multicolumn{1}{r|}{25.33\%}        & \multicolumn{1}{r|}{40.27\%}        & \multicolumn{1}{r|}{8.00\%}         & \multicolumn{1}{r|}{27.73\%}        & \multicolumn{1}{r|}{35.47\%}        & 14.67\%                             & 25.24\%                                            \\ \hline
\multirow{3}{*}{drawings}                             & ResNet                                    & \multicolumn{1}{r|}{20.00\%}        & \multicolumn{1}{r|}{4.00\%}         & \multicolumn{1}{r|}{53.60\%}        & \multicolumn{1}{r|}{43.33\%}        & \multicolumn{1}{r|}{6.40\%}         & 12.53\%                             & \textbf{23.31\%}                                   \\ \cline{2-9} 
                                                      & Moco                                      & \multicolumn{1}{r|}{50.67\%}        & \multicolumn{1}{r|}{20.00\%}        & \multicolumn{1}{r|}{3.20\%}         & \multicolumn{1}{r|}{4.80\%}         & \multicolumn{1}{r|}{7.20\%}         & 52.00\%                             & 22.98\%                                            \\ \cline{2-9} 
                                                      & CLIP                                      & \multicolumn{1}{r|}{9.33\%}         & \multicolumn{1}{r|}{0.00\%}         & \multicolumn{1}{r|}{36.00\%}        & \multicolumn{1}{r|}{0.00\%}         & \multicolumn{1}{r|}{20.00\%}        & 22.40\%                             & 14.62\%                                            \\ \hline
text                                                  & CLIP                                      & \multicolumn{1}{r|}{20.00\%}        & \multicolumn{1}{r|}{2.40\%}         & \multicolumn{1}{r|}{0.00\%}         & \multicolumn{1}{r|}{10.40\%}        & \multicolumn{1}{r|}{16.67\%}        & 15.20\%                             & 10.78\%                                            \\ \hline
\end{tabular}
    \caption{Top-25 success rate for experiment in Franka Kitchen domain (DGSD).}
    \label{tab:franka_exp3}
\end{table}

\begin{table}[]
\centering
\scriptsize
\begin{tabular}{|c|c|cccccc|c|}
\hline
\textbf{Task Specification } & \multirow{2}{*}{\textbf{model}} & \multicolumn{6}{c|}{\textbf{Task ID}}                                                                                                                                                                        & \multirow{2}{*}{\textbf{avg}} \\ \cline{3-8}
              \textbf{ modality}                                       &                                           & \multicolumn{1}{c|}{\textbf{knob3}} & \multicolumn{1}{c|}{\textbf{rdoor}} & \multicolumn{1}{c|}{\textbf{sdoor}} & \multicolumn{1}{c|}{\textbf{knob2}} & \multicolumn{1}{c|}{\textbf{ldoor}} & \textbf{micro} &                               \\ \hline
\multirow{3}{*}{same\_scene\_imgs}                    & ResNet                                    & \multicolumn{1}{c|}{12.00\%}        & \multicolumn{1}{c|}{100.00\%}       & \multicolumn{1}{c|}{0.00\%}         & \multicolumn{1}{c|}{0.00\%}         & \multicolumn{1}{c|}{100.00\%}       & 0.00\%         & 35.33\%                       \\ \cline{2-9} 
                                                      & Moco                                      & \multicolumn{1}{c|}{0.00\%}         & \multicolumn{1}{c|}{100.00\%}       & \multicolumn{1}{c|}{100.00\%}       & \multicolumn{1}{c|}{0.00\%}         & \multicolumn{1}{c|}{0.00\%}         & 0.00\%         & 33.33\%                       \\ \cline{2-9} 
                                                      & CLIP                                      & \multicolumn{1}{c|}{0.00\%}         & \multicolumn{1}{c|}{100.00\%}       & \multicolumn{1}{c|}{64.00\%}        & \multicolumn{1}{c|}{100.00\%}       & \multicolumn{1}{c|}{100.00\%}       & 100.00\%       & \textbf{77.33\%}              \\ \hline
\multirow{3}{*}{online\_imgs}                         & ResNet                                    & \multicolumn{1}{c|}{0.00\%}         & \multicolumn{1}{c|}{0.00\%}         & \multicolumn{1}{c|}{0.00\%}         & \multicolumn{1}{c|}{0.00\%}         & \multicolumn{1}{c|}{6.00\%}         & 0.00\%         & 1.00\%                        \\ \cline{2-9} 
                                                      & Moco                                      & \multicolumn{1}{c|}{0.00\%}         & \multicolumn{1}{c|}{0.00\%}         & \multicolumn{1}{c|}{0.00\%}         & \multicolumn{1}{c|}{0.00\%}         & \multicolumn{1}{c|}{0.00\%}         & 0.00\%         & 0.00\%                        \\ \cline{2-9} 
                                                      & CLIP                                      & \multicolumn{1}{c|}{0.00\%}         & \multicolumn{1}{c|}{0.00\%}         & \multicolumn{1}{c|}{54.00\%}        & \multicolumn{1}{c|}{4.00\%}         & \multicolumn{1}{c|}{100.00\%}       & 64.00\%        & \textbf{37.00\%}              \\ \hline
\multirow{3}{*}{drawings}                             & ResNet                                    & \multicolumn{1}{c|}{100.00\%}       & \multicolumn{1}{c|}{0.00\%}         & \multicolumn{1}{c|}{0.00\%}         & \multicolumn{1}{c|}{0.00\%}         & \multicolumn{1}{c|}{0.00\%}         & 18.00\%        & \textbf{19.67\%}              \\ \cline{2-9} 
                                                      & Moco                                      & \multicolumn{1}{c|}{0.00\%}         & \multicolumn{1}{c|}{0.00\%}         & \multicolumn{1}{c|}{64.00\%}        & \multicolumn{1}{c|}{0.00\%}         & \multicolumn{1}{c|}{0.00\%}         & 0.00\%         & 10.67\%                       \\ \cline{2-9} 
                                                      & CLIP                                      & \multicolumn{1}{c|}{0.00\%}         & \multicolumn{1}{c|}{0.00\%}         & \multicolumn{1}{c|}{0.00\%}         & \multicolumn{1}{c|}{0.00\%}         & \multicolumn{1}{c|}{100.00\%}       & 0.00\%         & 16.67\%                       \\ \hline
text                                                  & CLIP                                      & \multicolumn{1}{c|}{0.00\%}         & \multicolumn{1}{c|}{0.00\%}         & \multicolumn{1}{c|}{0.00\%}         & \multicolumn{1}{c|}{0.00\%}         & \multicolumn{1}{c|}{0.00\%}         & 0.00\%         & 0.00\%                        \\ \hline
\end{tabular}
\caption{Top-25 success rate for goal selection task in Franka kitchen domain (SGDD).}
\label{tab:franka_exp2}
\end{table}

\begin{table}[]
\centering
\scriptsize
\begin{tabular}{|c|c|c|c|c|c|c|c|c|}
\hline
\textbf{Task Specification } & \multirow{2}{*}{\textbf{model}} & \multicolumn{6}{c|}{\textbf{Task ID}}                                                                                                                                                                        & \multirow{2}{*}{\textbf{avg}} \\ \cline{3-8}
              \textbf{ modality}                                       &                                           & \multicolumn{1}{c|}{\textbf{knob3}} & \multicolumn{1}{c|}{\textbf{rdoor}} & \multicolumn{1}{c|}{\textbf{sdoor}} & \multicolumn{1}{c|}{\textbf{knob2}} & \multicolumn{1}{c|}{\textbf{ldoor}} & \textbf{micro} &                               \\ \hline

\multirow{3}{*}{same\_scene\_imgs} & ResNet          & 0.00\%                & 0.00\%                  & 0.00\%                  & 0.00\%                & 0.00\%                  & 0.00\%                  & 0.00\%  \\ \cline{2-9} 
                                   & Moco            & 0.00\%                & 0.00\%                  & 0.00\%                  & 0.00\%                & 0.00\%                  & 0.00\%                  & 0.00\%  \\ \cline{2-9} 
                                   & CLIP            & 0.00\%                & 24.00\%                 & 0.00\%                  & 33.33\%               & 33.33\%                 & 0.00\%                  & 15.11\% \\ \hline
\multirow{3}{*}{online\_imgs}      & ResNet          & 6.00\%                & 0.00\%                  & 11.56\%                 & 0.00\%                & 0.00\%                  & 0.00\%                  & 2.93\%  \\ \cline{2-9} 
                                   & Moco            & 0.00\%                & 32.89\%                 & 0.00\%                  & 0.00\%                & 14.00\%                 & 0.00\%                  & 7.81\%  \\ \cline{2-9} 
                                   & CLIP            & 12.00\%               & 33.33\%                 & 0.00\%                  & 0.00\%                & 0.00\%                  & 0.00\%                  & 7.56\%  \\ \hline
\multirow{3}{*}{drawings}          & ResNet          & 0.00\%                & 0.00\%                  & 0.00\%                  & 0.00\%                & 0.00\%                  & 0.00\%                  & 0.00\%  \\ \cline{2-9} 
                                   & Moco            & 0.00\%                & 33.33\%                 & 0.00\%                  & 0.00\%                & 0.00\%                  & 0.00\%                  & 5.56\%  \\ \cline{2-9} 
                                   & CLIP            & 18.22\%               & 0.00\%                  & 30.67\%                 & 0.00\%                & 32.89\%                 & 0.00\%                  & 13.63\% \\ \hline
text                               & CLIP            & 0.00\%                & 0.00\%                  & 2.67\%                  & 33.33\%               & 0.00\%                  & 0.00\%                  & 6.00\%  \\ \hline
\end{tabular}
\caption{Top-25 success rate for goal selection task in Franka kitchen domain (DGDD).}
\label{tab:franka_exp4}
\end{table}

\begin{table}[]
    \centering
    \scriptsize
   \begin{tabular}{|c|c|rrrrrr|r|}
\hline
\textbf{Task Specification } & \multirow{2}{*}{\textbf{model}} & \multicolumn{6}{c|}{\textbf{Task ID}}                                                                                                                                                                                             & \multicolumn{1}{c|}{\multirow{2}{*}{\textbf{avg}}} \\ \cline{3-8}
      \textbf{modality}                                                &                                           & \multicolumn{1}{c|}{\textbf{knob3}} & \multicolumn{1}{c|}{\textbf{rdoor}} & \multicolumn{1}{c|}{\textbf{sdoor}} & \multicolumn{1}{c|}{\textbf{knob2}} & \multicolumn{1}{c|}{\textbf{ldoor}} & \multicolumn{1}{c|}{\textbf{micro}} & \multicolumn{1}{c|}{}                              \\ \hline
\multirow{3}{*}{same\_scene\_imgs}                    & ResNet                                    & \multicolumn{1}{r|}{16.00\%}        & \multicolumn{1}{r|}{100.00\%}       & \multicolumn{1}{r|}{96.00\%}        & \multicolumn{1}{r|}{100.00\%}       & \multicolumn{1}{r|}{100.00\%}       & 96.00\%                             & 84.67\%                                            \\ \cline{2-9} 
                                                      & Moco                                      & \multicolumn{1}{r|}{56.00\%}        & \multicolumn{1}{r|}{100.00\%}       & \multicolumn{1}{r|}{100.00\%}       & \multicolumn{1}{r|}{100.00\%}       & \multicolumn{1}{r|}{100.00\%}       & 100.00\%                            & 92.67\%                                            \\ \cline{2-9} 
                                                      & CLIP                                      & \multicolumn{1}{r|}{100.00\%}       & \multicolumn{1}{r|}{100.00\%}       & \multicolumn{1}{r|}{64.00\%}        & \multicolumn{1}{r|}{100.00\%}       & \multicolumn{1}{r|}{100.00\%}       & 100.00\%                            & \textbf{94.00\%}                                   \\ \hline
\multirow{3}{*}{online\_imgs}                         & ResNet                                    & \multicolumn{1}{r|}{32.80\%}        & \multicolumn{1}{r|}{0.00\%}         & \multicolumn{1}{r|}{100.00\%}       & \multicolumn{1}{r|}{0.00\%}         & \multicolumn{1}{r|}{54.40\%}        & 32.80\%                             & 36.67\%                                            \\ \cline{2-9} 
                                                      & Moco                                      & \multicolumn{1}{r|}{60.00\%}        & \multicolumn{1}{r|}{100.00\%}       & \multicolumn{1}{r|}{21.60\%}        & \multicolumn{1}{r|}{61.60\%}        & \multicolumn{1}{r|}{71.20\%}        & 52.00\%                             & 61.07\%                                            \\ \cline{2-9} 
                                                      & CLIP                                      & \multicolumn{1}{r|}{18.40\%}        & \multicolumn{1}{r|}{98.40\%}        & \multicolumn{1}{r|}{68.00\%}        & \multicolumn{1}{r|}{8.80\%}         & \multicolumn{1}{r|}{100.00\%}       & 89.60\%                             & \textbf{63.87\%}                                   \\ \hline
\multirow{3}{*}{drawings}                             & ResNet                                    & \multicolumn{1}{r|}{100.00\%}       & \multicolumn{1}{r|}{0.00\%}         & \multicolumn{1}{r|}{0.00\%}         & \multicolumn{1}{r|}{8.00\%}         & \multicolumn{1}{r|}{0.00\%}         & 14.40\%                             & 20.40\%                                            \\ \cline{2-9} 
                                                      & Moco                                      & \multicolumn{1}{r|}{0.00\%}         & \multicolumn{1}{r|}{100.00\%}       & \multicolumn{1}{r|}{33.60\%}        & \multicolumn{1}{r|}{0.00\%}         & \multicolumn{1}{r|}{0.00\%}         & 4.00\%                              & 22.93\%                                            \\ \cline{2-9} 
                                                      & CLIP                                      & \multicolumn{1}{r|}{38.40\%}        & \multicolumn{1}{r|}{0.00\%}         & \multicolumn{1}{r|}{96.00\%}        & \multicolumn{1}{r|}{0.00\%}         & \multicolumn{1}{r|}{100.00\%}       & 76.00\%                             & \textbf{51.73\%}                                   \\ \hline
text                                                  & CLIP                                      & \multicolumn{1}{r|}{0.00\%}         & \multicolumn{1}{r|}{0.00\%}         & \multicolumn{1}{r|}{8.00\%}         & \multicolumn{1}{r|}{100.00\%}       & \multicolumn{1}{r|}{0.00\%}         & 0.00\%                              & 18.00\%                                            \\ \hline
\end{tabular}
    \caption{Action Selection task for Franka Kitchen Domain}
    \label{tab:franka_action_selection}
\end{table}
}

\section{Details of Goal Selection Experiments in Franka Kitchen Domain}
\begin{enumerate}
 \item \textbf{SGSD}: As shown in Table~\ref{tab:franka_kitchen}, CLIP outperform other embedding models with image-based task specification modalities using \textbf{Cos+Delta}.
    
    \item \textbf{DGSD}:  As shown in Table~\ref{tab:franka_exp3}, CLIP does not outperform the other two embedding models.
    
    \item \textbf{SGDD}: As shown in Table~\ref{tab:franka_exp2}, CLIP outperforms the other two models under 2 out of the 3 image-based task specification modalities in the goal selection task in the Franka kitchen domain (using \textbf{Cos+Delta}).
    
    \item \textbf{DGDD}: As shown in Table~\ref{tab:franka_exp4}, CLIP outperforms the other two models in this hardest setting as well. However, the overall performance is not very good.
    
    \item \textbf{ASDD}(Action Selection):  As shown in Table~\ref{tab:franka_action_selection}, CLIP outperforms the other two embedding models in this scenario.

\end{enumerate}

{\renewcommand{\arraystretch}{1.2}

\begin{table}[]
\centering
\scriptsize
\begin{tabular}{|c|c|c|c|c|c|c|c|c|c|}
\hline
\multirow{3}{*}{\textbf{Exp}} &  \multirow{3}{*}{\textbf{Model}} &   \multicolumn{7}{c|}{\textbf{Task ID}} & \multirow{3}{*}{All tasks}  \\ \cline{3-9}
                                         &  &  \multirow{2}{*}{door}  &  \multirow{2}{*}{drawer} &  \multirow{2}{*}{box} &  \multirow{2}{*}{fridge} & \multirow{2}{*}{laptop} &  washing &  \multirow{2}{*}{oven} & \\ 
                                         & & & & & & & machine & & \\ \hline
\multirow{3}{*}{real2real} & CLIP           & 38.00\%       & 46.00\%         & 61.00\%      & 38.00\%               & 23.00\%         & 55.00\%                  & 69.00\%       & 47.14\%          \\ \cline{2-10} 
                           & ResNet         & 96.00\%       & 99.00\%         & 68.00\%      & 100.00\%              & 97.00\%         & 68.00\%                  & 100.00\%      & \textbf{89.71\%} \\ \cline{2-10} 
                           & Moco           & 88.00\%       & 94.00\%         & 66.00\%      & 100.00\%              & 95.00\%         & 84.00\%                  & 100.00\%      & 89.57\%          \\ \hline
\multirow{3}{*}{sim2real}  & CLIP           & 46.00\%       & 46.00\%         & 53.00\%      & 42.00\%               & 34.00\%         & 40.00\%                  & 81.00\%       & 48.86\%          \\ \cline{2-10} 
                           & ResNet         & 61.00\%       & 51.00\%         & 32.00\%      & 36.00\%               & 98.00\%         & 9.00\%                   & 77.00\%       & 52.00\%          \\ \cline{2-10} 
                           & Moco           & 35.00\%       & 90.00\%         & 36.00\%      & 68.00\%               & 94.00\%         & 54.00\%                  & 70.00\%       & \textbf{63.86\%} \\ \hline
\end{tabular}
\caption{Top-25 success rate for experiment with SSV2 dataset: SGSD}
\label{tab:sapien_SGSD}
\end{table}

\begin{table}[]
\centering
\scriptsize
\begin{tabular}{|c|c|c|c|c|c|c|c|c|c|}
\hline
\multirow{3}{*}{\textbf{Exp}} &  \multirow{3}{*}{\textbf{Model}} &   \multicolumn{7}{c|}{\textbf{Task ID}} & \multirow{3}{*}{All tasks}  \\ \cline{3-9}
                                         &  &  \multirow{2}{*}{door}  &  \multirow{2}{*}{drawer} &  \multirow{2}{*}{box} &  \multirow{2}{*}{fridge} & \multirow{2}{*}{laptop} &  washing &  \multirow{2}{*}{oven} & \\ 
                                         & & & & & & & machine & & \\ \hline
\multirow{3}{*}{real2real} & CLIP           & 46.00\%                & 40.00\%                  & 42.00\%                  & 42.00\%                        & 44.00\%                  & 48.00\%                            & 44.00\%                & 43.71\%                               \\ \cline{2-10} 
                           & ResNet         & 50.00\%                & 59.00\%                  & 64.00\%                  & 55.00\%                        & 31.00\%                  & 49.00\%                            & 56.00\%                & \textbf{52.00\%}                      \\ \cline{2-10} 
                           & Moco           & 60.00\%                & 56.00\%                  & 51.00\%                  & 52.00\%                        & 34.00\%                  & 55.00\%                            & 49.00\%                & 51.00\%                               \\ \hline
\multirow{3}{*}{sim2real}  & CLIP           & 42.00\%                & 40.00\%                  & 52.00\%                  & 41.00\%                        & 47.00\%                  & 46.00\%                            & 43.00\%                & 44.43\%                               \\ \cline{2-10} 
                           & ResNet         & 35.00\%                & 58.00\%                  & 42.00\%                  & 58.00\%                        & 37.00\%                  & 58.00\%                            & 41.00\%                & 47.00\%                               \\ \cline{2-10} 
                           & Moco           & 64.00\%                & 61.00\%                  & 50.00\%                  & 68.00\%                        & 23.00\%                  & 52.00\%                            & 67.00\%                & \textbf{55.00\%}                      \\ \hline
\end{tabular}
\caption{Top-25 success rate for experiment with SSV2 dataset: DGSD}
\label{tab:sapien_DGSD}
\end{table}

\begin{table}[]
\centering
\scriptsize
\begin{tabular}{|c|c|c|c|c|c|c|c|c|c|}
\hline
\multirow{3}{*}{\textbf{Exp}} &  \multirow{3}{*}{\textbf{Model}} &   \multicolumn{7}{c|}{\textbf{Task ID}} & \multirow{3}{*}{All tasks}  \\ \cline{3-9}
                                         &  &  \multirow{2}{*}{door}  &  \multirow{2}{*}{drawer} &  \multirow{2}{*}{box} &  \multirow{2}{*}{fridge} & \multirow{2}{*}{laptop} &  washing &  \multirow{2}{*}{oven} & \\ 
                                         & & & & & & & machine & & \\ \hline
\multirow{3}{*}{real2real} & CLIP           & 12.00\%                & 8.00\%                   & 1.00\%                   & 4.00\%                         & 1.00\%                   & 11.00\%                            & 5.00\%                 & 6.00\%                         \\ \cline{2-10} 
                           & ResNet         & 21.00\%                & 39.00\%                  & 0.00\%                   & 100.00\%                       & 92.00\%                  & 25.00\%                            & 67.00\%                & 49.14\%               \\ \cline{2-10} 
                           & Moco           & 0.00\%                 & 57.00\%                  & 1.00\%                   & 100.00\%                       & 100.00\%                 & 25.00\%                            & 100.00\%               & \textbf{54.71\%}               \\ \hline
\multirow{3}{*}{sim2real}  & CLIP           & 1.00\%                 & 1.00\%                   & 11.00\%                  & 5.00\%                         & 1.00\%                   & 12.00\%                            & 7.00\%                 & 5.43\%                         \\ \cline{2-10} 
                           & ResNet         & 8.00\%                 & 29.00\%                  & 20.00\%                  & 0.00\%                         & 97.00\%                  & 0.00\%                             & 11.00\%                & 23.57\%              \\ \cline{2-10} 
                           & Moco           & 33.00\%                & 52.00\%                  & 23.00\%                  & 0.00\%                         & 92.00\%                  & 0.00\%                             & 0.00\%                 & \textbf{28.57\%}               \\ \hline

\end{tabular}
\caption{Top-25 success rate for experiment with SSV2 dataset: SGDD}
\label{tab:sapien_SGDD}
\end{table}

\begin{table}[]
\centering
\scriptsize
\begin{tabular}{|c|c|c|c|c|c|c|c|c|c|}
\hline
\multirow{3}{*}{\textbf{Exp}} &  \multirow{3}{*}{\textbf{Model}} &   \multicolumn{7}{c|}{\textbf{Task ID}} & \multirow{3}{*}{All tasks}  \\ \cline{3-9}
                                         &  &  \multirow{2}{*}{door}  &  \multirow{2}{*}{drawer} &  \multirow{2}{*}{box} &  \multirow{2}{*}{fridge} & \multirow{2}{*}{laptop} &  washing &  \multirow{2}{*}{oven} & \\ 
                                         & & & & & & & machine & & \\ \hline
\multirow{3}{*}{real2real} & CLIP           & 6.00\%                 & 4.00\%                   & 6.00\%                   & 6.00\%                         & 3.00\%                   & 6.00\%                             & 7.00\%                 & 5.43\%                         \\ \cline{2-10} 
                           & ResNet         & 9.00\%                 & 2.00\%                   & 8.00\%                   & 0.00\%                         & 0.00\%                   & 1.00\%                             & 4.00\%                 & 3.43\%                \\ \cline{2-10} 
                           & Moco           & 15.00\%                & 13.00\%                  & 16.00\%                  & 0.00\%                         & 0.00\%                   & 4.00\%                             & 0.00\%                 & \textbf{6.86\%}                \\ \hline
\multirow{3}{*}{sim2real}  & CLIP           & 8.00\%                 & 6.00\%                   & 6.00\%                   & 11.00\%                        & 3.00\%                   & 8.00\%                             & 5.00\%                 & 6.71\%                         \\ \cline{2-10} 
                           & ResNet         & 12.00\%                & 7.00\%                   & 9.00\%                   & 9.00\%                         & 0.00\%                   & 12.00\%                            & 0.00\%                 & 7.00\%             \\ \cline{2-10} 
                           & Moco           & 7.00\%                 & 15.00\%                  & 7.00\%                   & 5.00\%                         & 0.00\%                   & 17.00\%                            & 4.00\%                 & \textbf{7.86\%}                \\ \hline
\end{tabular}
\caption{Top-25 success rate for experiment with SSV2 dataset: DGDD}
\label{tab:sapien_DGDD}
\end{table}

\begin{table}[]
    \centering
    \scriptsize
    \begin{tabular}{|c|c|c|c|c|c|c|c|c|c|}
        \hline 
    
     \multirow{3}{*}{\textbf{Exp}} &  \multirow{3}{*}{\textbf{Model}} &   \multicolumn{7}{c|}{\textbf{Task ID}} & \multirow{3}{*}{All tasks}  \\ \cline{3-9}
                                         &  &  \multirow{2}{*}{door}  &  \multirow{2}{*}{drawer} &  \multirow{2}{*}{box} &  \multirow{2}{*}{fridge} & \multirow{2}{*}{laptop} &  washing &  \multirow{2}{*}{oven} & \\ 
                                         & & & & & & & machine & & \\ \hline
     \multirow{2}{*}{R2R} &   CLIP  & 0.65 & 0.64 & 0.28 & 1.0 & 1.0 & 0.64 & 1.0 & 0.74 \\\cline{2-10}
                             & ResNet  & 0.63 & 0.75 & 0.62 & 0.94 & 1.0 & 0.78 & 0.8 & 0.79 \\ \hline
        
        \multirow{2}{*}{S2S} &   CLIP  & 0.75 & 0.23 & 0.14 & 0.66 & 0.37 & 0.72 & 0.93 & 0.54 \\\cline{2-10}
                             & ResNet  & 0.46 & 0.42 & 0.98 & 0.97 & 0.81 & 0.7 & 1.0 & 0.76 \\ \hline
       
      \multirow{2}{*}{R2S} &   CLIP   & 0.2 & 0.32 & 0.2 & 0.0 & 0.13 & 0.13 & 0.24 & 0.17 \\\cline{2-10}
                            & ResNet   & 	0.0 & 0.1 & 0.21 & 0.03 & 0.93 & 0.11 & 0.29 & 0.24 \\ \hline
       
      \multirow{2}{*}{S2R} &  CLIP    & 0.18 & 0.35 & 0.1 & 0.54 & 0.78 & 0.06 & 0.3 & 0.33  \\\cline{2-10}
                            & ResNet   & 1.0 & 0.97 & 0.0 & 0.4 & 0.8 & 0.28 & 0.62 & 0.58 \\ \hline

    \end{tabular}
    \caption{ Top-25 success rate for experiment with SSV2 dataset: ASDD}
    \label{tab:action_selection}
\end{table}

}

\section{Details of Goal Selection Experiments in SSV2 Dataset}

For this experiment, we test \textbf{Cos+Delta} with three different embedding models. We conducted each experiment with 10 different repetitions, each time sampling a different trajectory for specifying the task. 
The results for the five different task scenarios SGSD,DGSD,SGDD,DGDD, and ASDD are shown in Table~\ref{tab:sapien_SGSD},\ref{tab:sapien_DGSD},\ref{tab:sapien_SGDD},\ref{tab:sapien_DGDD}
and \ref{tab:action_selection} respectively. As shown, the performance of CLIP is much lower than that of ResNet and Moco with SSV2 dataset.

\section{Goal Similarity Scores for Multi-step Tasks}

\begin{figure}
    \centering
    \includegraphics[width=0.92\linewidth]{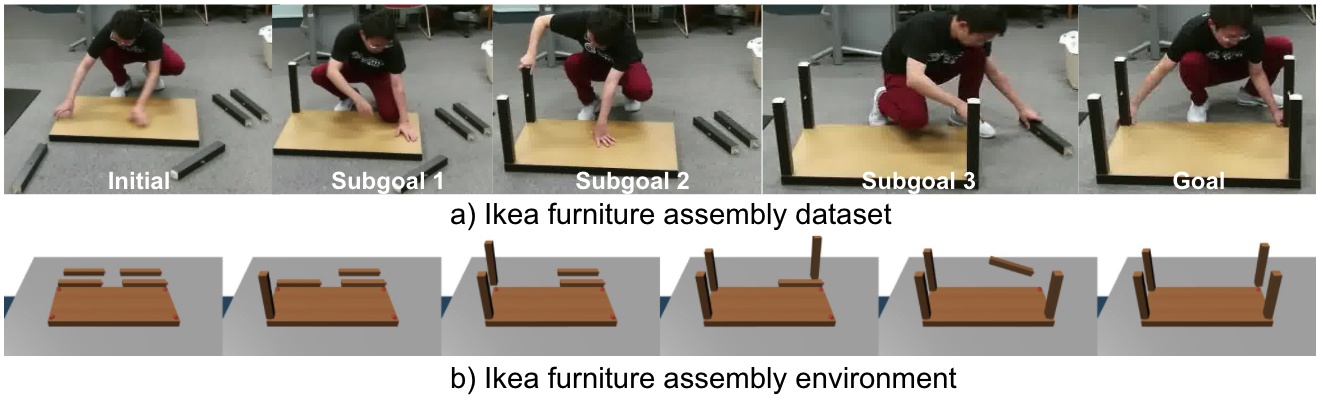}
    \vspace{-0.3cm}
    \caption{Sample image frames from IKEA furniture assembly dataset and simulation environment.}
    \label{fig:furniture}
\end{figure}

\begin{figure}
    \centering
    \includegraphics[width=0.92\linewidth]{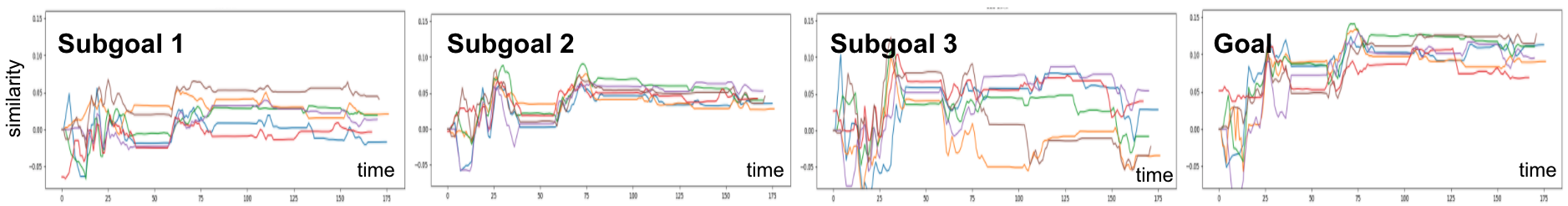}
    \vspace{-0.3cm}
    \caption{Cosine similarities (from the CLIP-based model) to each subgoal for simulated furniture assembly trajectories.}
    \label{fig:furniture_rewards}
\end{figure}

%furniture assembly dataset and ikea assembly simulation
Many realworld tasks are multi-step and consist of different numbers of sub-goals. 
In this experiment, we evaluate how our proposed method work for multi-step tasks and investigate whether specifying sub-goals improves its performance.
More specifically, we study the task of table assembly with real world video data from the IKEA furniture assembly dataset~\citep{ikeadataset} and synthetic visual trajectories in the IKEA assembly environment~\citep{lee2021ikea}. Sample trajectories from the dataset and the simulator are shown in Figure~\ref{fig:furniture}. This setting corresponds to when an agent is learning from human demonstrations in the form of video recordings and it has access to past experience data in simulation. 
Figure~\ref{fig:furniture_rewards} shows the consine similarities to each subgoal for different simulated trajectories. The average similarity across time increases from the first subgoal to the final goal since the final goal contains the subgoals. However, the similarity score pattern is not clear enough for telling apart at which point a subgoal is achieved. A more complex method is needed to find subgoals in these trajectories.
%We adapt our proposed method to incorporate multiple sub-goals by aggregating the similarity scores to all goal state vectors. We evaluate and compare the goal selection success rates with and without subgoals in this experiment.

\section{T-REX Training}

\begin{table}[]
    \centering
\vspace{0.1cm}
\scriptsize
{\renewcommand{\arraystretch}{1.2}
\begin{tabular}{ccccc}
\cline{1-4}
\multicolumn{1}{|l|}{}      & \multicolumn{1}{c|}{ResNet}  & \multicolumn{1}{c|}{Moco}    & \multicolumn{1}{c|}{CLIP}    &    \\ \cline{1-4}
\multicolumn{1}{|c|}{knob2} & \multicolumn{1}{c|}{71.71\%} & \multicolumn{1}{c|}{69.51\%} & \multicolumn{1}{c|}{72.20\%} &    \\ \cline{1-4}
\multicolumn{1}{|c|}{knob3} & \multicolumn{1}{c|}{76.10\%} & \multicolumn{1}{c|}{69.51\%} & \multicolumn{1}{c|}{80.49\%} &    \\ \cline{1-4}
\multicolumn{1}{|c|}{ldoor} & \multicolumn{1}{c|}{76.34\%} & \multicolumn{1}{c|}{78.05\%} & \multicolumn{1}{c|}{76.34\%} &    \\ \cline{1-4}
\multicolumn{1}{|c|}{micro} & \multicolumn{1}{c|}{84.88\%} & \multicolumn{1}{c|}{85.12\%} & \multicolumn{1}{c|}{86.34\%} &   \\ \cline{1-4}
\multicolumn{1}{|c|}{rdoor} & \multicolumn{1}{c|}{71.71\%} & \multicolumn{1}{c|}{71.71\%} & \multicolumn{1}{c|}{71.71\%} &    \\ \cline{1-4}
\multicolumn{1}{|c|}{sdoor} & \multicolumn{1}{c|}{86.34\%} & \multicolumn{1}{c|}{84.88\%} & \multicolumn{1}{c|}{80.49\%} &    \\ \cline{1-4}
\end{tabular}
}
 \caption{Max. T-REX test accuracy under each model.}
\label{table:trex_max_acc}
\end{table}

\begin{figure}
    \centering
    \includegraphics[width=\textwidth]{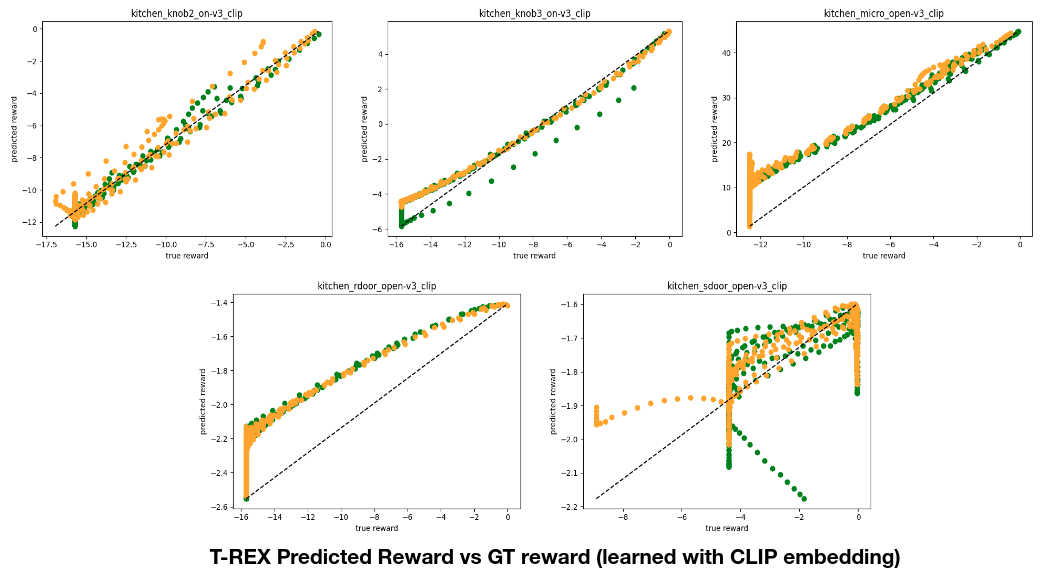}
    \caption{Return predicted by T-REX models vs GT returns. Green dots are seen trajectories that were used in training and yellow dots are unseen trajectories.}
    \label{fig:trex_rewards}
\end{figure}

\begin{figure}
    \centering
    \includegraphics[width=0.6\textwidth]{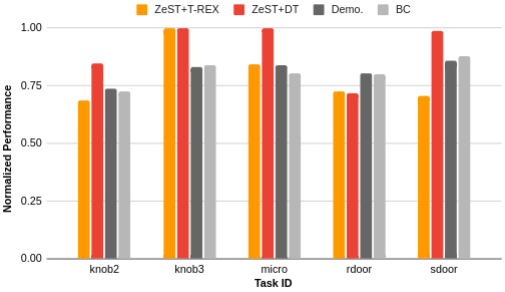}
    \caption{Performance T-REX policies comparing with other methods}
    \label{fig:trex_results}
\end{figure}

We also employ a ranking-based reward learning method, T-REX~\citep{brown2019extrapolating}, to learn a reward function. 
% %T-REX learns a reward function in a supervised way by enforcing the sum of predicted return for a pair of trajectories match their relative ranking. 
We train T-REX with trajectories ranked by aggregated cosine similarities. We then test on classifying sub-trajectories (relative ranking) with ground-truth dense reward rankings. 
Table~\ref{table:trex_max_acc} shows the maximum test accuracy under each model for different tasks. 
Figure~\ref{fig:trex_rewards} shows the relationship between predicted returns and ground truth returns on both seen and unseen trajectories. The return predicted by T-REX models correlate well with ground truth return for four out of the five tasks.
We observe that the predicted rewards correlate well with ground-truth rewards.
We then use the learned rewards to train policy gradient agents. 
Figure~\ref{fig:trex_results} shows the learned policies' performance with different methods under each task. The return is normalized to the expert's performance as 1.0. 
T-REX training is able to generate policies that outperform behavioral cloning in 2 out of the 5 tasks, while T-REX does not need the action labels from the demonstrations. 
% Decision transformer policies outperform baseline in 3 out of the 5 tasks, indicating ZeST scores provide additional information for the learning agent to extrapolate from suboptimal demonstrations.
% %T-REX training improved the learned policies' performance on 3 of the 5 tasks with 1 task worse than using raw similarity scores as rewards.

\end{document}